\documentclass{article}

    \usepackage[final]{neurips_2025}

\usepackage[utf8]{inputenc} 
\usepackage[T1]{fontenc}    
\usepackage{hyperref}       
\usepackage{url}            
\usepackage{booktabs}       
\usepackage{amsfonts}       
\usepackage{nicefrac}       
\usepackage{microtype}      
\usepackage{xcolor}         
\usepackage{enumitem} 
\usepackage{amsmath,amssymb,amsthm}
\usepackage{siunitx}
\usepackage{graphicx}
\usepackage{subcaption}

\usepackage{algorithm}
\usepackage{algorithmic}

\newtheorem{theorem}{Theorem}
\newtheorem{lemma}{Lemma}
\title{Bootstrap Off-policy with World Model}

\author{
Guojian Zhan$^{1,2}$, \quad
Likun Wang$^{1}$, \quad
Xiangteng Zhang$^{1}$, \quad
Jiaxin Gao$^{1}$, \\
\textbf{Masayoshi Tomizuka}$^{2}$,\quad
\textbf{Shengbo Eben Li}$^{1}$\thanks{Corresponding author. \texttt{lishbo@tsinghua.edu.cn.}} \\
\vspace{0.15cm}\\
$^1$ College of AI \& School of Vehicle and Mobility, Tsinghua University \\ $^2$ Berkeley AI Research (BAIR), UC Berkeley
}

\begin{document}

\maketitle

\begin{abstract}
Online planning has proven effective in reinforcement learning (RL) for improving sample efficiency and final performance. However, using planning for environment interaction inevitably introduces a divergence between the collected data and the policy's actual behaviors, degrading both model learning and policy improvement. To address this, we propose BOOM (Bootstrap Off-policy with WOrld Model), a framework that tightly integrates planning and off-policy learning through a bootstrap loop: the policy initializes the planner, and the planner refines actions to bootstrap the policy through behavior alignment. This loop is supported by a jointly learned world model, which enables the planner to simulate future trajectories and provides value targets to facilitate policy improvement. The core of BOOM is a likelihood-free alignment loss that bootstraps the policy using the planner’s non-parametric action distribution, combined with a soft value-weighted mechanism that prioritizes high-return behaviors and mitigates variability in the planner’s action quality within the replay buffer. Experiments on the high-dimensional DeepMind Control Suite and Humanoid-Bench show that BOOM achieves state-of-the-art results in both training stability and final performance. The code is accessible at \url{https://github.com/molumitu/BOOM_MBRL}.
\end{abstract}

\section{Introduction}
Reinforcement learning (RL) has achieved impressive performance in a wide range of domains, from industrial automation to autonomous driving and embodied intelligence~\cite{vinyals2019grandmaster, schrittwieser2020mastering, zhan2024transformation}. Among the various techniques developed to enhance RL, online planning stands out for its predictive optimization ability to improve control performance using learned dynamics~\cite{moerland2023model, campbell2019model, zhan2023enhance}. By performing look-ahead rollouts, it enables agents to anticipate future consequences and iteratively refine actions~\cite{tassa2012synthesis, schrittwieser2021online}. Compared to model-free approaches, which rely solely on trial-and-error learning, model-based planning offers an effective tool to generate high-quality actions for environment interaction~\cite{deisenroth2011pilco, curi2020efficient}.


A growing body of research has explored how to more effectively integrate planning into RL~\cite{hamrickrole, moerland2023model}. Early methods such as PETS~\cite{chua2018deep} and PlaNet~\cite{hafner2019learning} have proven that planning with learned dynamics and reward signals can achieve impressive control performance. 
To further improve performance in high-dimensional tasks, recent approaches have combined online planning with policy learning, where the policy can provide a good initial solution to speed up planning~\cite{argenson2020model}. For example, 
LOOP~\cite{sikchi2022learning} builds on SAC~\cite{haarnoja2018soft}, an off-policy model-free algorithm, and integrates a planner guided by a learned dynamics model to enable higher-quality environment interaction. Recently, TD-MPC and TD-MPC2~\cite{hansen2022temporal,hansen2023td} jointly learn the dynamics, reward and value functions through temporal differential (TD) learning, achieving strong performance through both algorithmic innovations and implementation advances. These methods successfully deliver substantial gains over model-free baselines such as SAC, particularly on high-dimensional benchmarks.


However, this class of planning-driven model-based RL algorithms inevitably suffers from a fundamental issue known as \textit{actor divergence}: the data used for learning is collected by the planner, which acts as a different actor from the policy network~\cite{sikchi2022learning}. Under the paradigm framework of off-policy RL, this issue leads to two problems: (1) \textit{Distribution shift in value learning}: The value function is trained on the data collected by the planner rather than the policy itself. As a result, it learns accurately within the planner's state-action distribution but tends to overestimate values in out-of-distribution regions that are rarely visited~\cite{kumar2020conservative}. (2) \textit{Unreliable policy improvement}: The policy network is updated using value estimates from the value network, which are influenced by the distributional shift. These biased estimates may mislead the policy, impairing its ability to distinguish between good and bad actions, ultimately leading to severe performance degradation~\cite{prudencio2023survey}.
To conclude, the misalignment between the training data and the policy's actual behavior can severely hinder the learning process, particularly in complex, high-dimensional environments where accurate value estimation is already challenging~\cite{nauman2024overestimation}. The resulting value bias not only misguides policy updates but also risks destabilizing the entire training process~\cite{markowitz2024avoiding}.
More critically, since online planning algorithms typically rely on sample-based optimization, the resulting action distributions are non-parametric and difficult to access. This means that they cannot be explicitly represented and the likelihood is intractable~\cite{williams2017model}. 

To address these challenges, we propose \textbf{B}ootstrap \textbf{O}ff-policy with W\textbf{O}rld \textbf{M}odel (BOOM), a novel framework that seamlessly integrates online planning with off-policy RL, effectively mitigating the negative impact of data distribution shifts caused by actor divergence.
This is accomplished through a bootstrap loop: the policy initializes the planner, and the planner refines actions to bootstrap the policy via behavior alignment, alleviating the \textit{actor divergence} issue. This loop is supported by a jointly learned world model, which enables the planner to simulate future trajectories and provides value targets that facilitate policy improvement.
We refer to it as bootstrap alignment because the planner typically generates higher-quality actions via model predictive optimization. Aligning the policy with these actions also offers strong guidance for improvement and accelerates learning.

In this paper, we introduce three key contributions:  \textbf{(1)} To facilitate alignment with the online planner's sample-based non-parametric distribution, we adopt a \textit{likelihood-free alignment loss} that measures the divergence between the policy and the planner without requiring explicit likelihoods of actions from the online planner. \textbf{(2)} 
We introduce a \textit{soft value-weighted mechanism} that prioritizes high-return behaviors, driven by the planner's value-guided action selection principle. Additionally, to maintain training efficiency, we align the policy with the stored planner actions in the replay buffer. Since these historical actions may vary in quality, our value-weighting mechanism ensures the policy prioritizes the high-valued promising experiences, accelerating learning while handling the variability in the planner’s past actions.
\textbf{(3)} BOOM combines ease of implementation with state-of-the-art (SOTA) performance on high-dimensional continuous control benchmarks, including the DeepMind Control Suite~\cite{tassa2018deepmind} and Humanoid-Bench~\cite{sferrazza2024humanoidbench}.

\section{Preliminaries}

\subsection{Reinforcement Learning}
Reinforcement learning (RL) offers a powerful framework for sequential decision-making~\cite{li2023rlbook}, formalized as a Markov Decision Process (MDP) \((\mathcal{S}, \mathcal{A}, P, r, \gamma)\), where \(\mathcal{S}\) and \(\mathcal{A}\) denote the state and action spaces, \(P(s'|s,a)\) is the transition dynamics, \(r(s,a)\) is the reward function, and \(\gamma \in [0,1)\) is the discount factor. At each timestep, an agent observes state \(s_t\), selects action \(a_t\), receives reward \(r_t\), and transitions to the next state \(s_{t+1}\) according to \(P\).
A fundamental concept in RL is the action-value function:
$
Q^\pi(s,a) = \mathbb{E}_\pi \left[ \sum_{t=0}^\infty \gamma^t r(s_t, a_t) \,\big|\, s_0 = s, a_0 = a \right],
$
which captures the expected return of taking action \(a\) in state \(s\), followed by policy \(\pi\).
Off-policy RL aims to discover the optimal policy \(\pi^*\) by maximizing the Q-value function using transitions generated from a different behavior policy \(\beta\). By decoupling data collection from policy improvement, off-policy methods can achieve high sample efficiency—primarily due to the effective reuse of past transitions stored in a replay buffer. However, when the behavior policy \(\beta\) deviates too far from \(\pi\), the resulting \textit{distributional shift} can severely undermine value estimation, a delicate yet critical challenge that continues to motivate much of the recent progress in off-policy learning~\cite{DSACT}.

\subsection{Online Planner}
Online planning optimizes action sequences at each step by simulating future trajectories under a predictive model, enabling informed action selection~\cite{zhan2024explicit}. This approach typically yields high-quality decisions by leveraging foresight over potential future outcomes. As a result, it is widely adopted in model-based RL for environment interaction, generating higher-quality interaction samples that facilitate more efficient learning and policy improvement.

Among various planning methods, {Model Predictive Path Integral} (MPPI) is a widely adopted sampling-based planner for continuous control tasks due to its high efficiency~\cite{williams2017model}. 
Formally, at each planning step, with the help of a dynamics model, $N_p$ action trajectories are sampled from a factorized Gaussian:
$
a_t^{i} \sim \mathcal{N}(\mu_t, \sigma_t^2), \quad \text{for } t = 0, \dots, H{-}1 \text{ and } i = 1, \dots, N_p,
$ where $H$ is the predictive horizon.
Each trajectory is evaluated using a reward and value model to estimate the future return:
$
G^{i} = \sum_{t=0}^{H-1} r_t^{i} + \hat{v}_H^{i},
$
where $r_t^{i}$ denotes predicted rewards and $\hat{v}_H^{i}$ is the terminal value estimate.
To update the trajectory distribution, MPPI reweights the $N$ samples using a softmax over the returns:
$
w^{i} = {\exp(G^{i} - \max_j G^{j})}/{\sum_k \exp(G^{k} - \max_j G^{j})}.
$
Next, the parameters of the trajectory distribution are updated using weighted statistics:
$
\mu_t \leftarrow \sum_i w^{i} a_t^{i}.
$
This process is repeated over several iterations, gradually converging toward an optimal solution. At test time, the planner executes the first action $ \mu_0$, while during training, Gaussian noise is added for exploration.

Despite its effectiveness, MPPI and similar planners suffer from a key limitation: the resulting planned action is obtained via weighted averaging over sampled candidates and does not directly arise from a parameterized probabilistic policy. 
Although Gaussian noise is used during the sampling stage, the final action distribution no longer adheres to a true Gaussian form because of the reweighting and resampling process. As a result, computing the precise likelihood of MPPI-generated actions is intractable in practice.

\section{Method}

\subsection{Inevitable Actor Divergence When Off-policy RL Meets Online Planning}

A common approach to integrating online planning with off-policy RL is to jointly learn a latent world model and a policy from replay buffer data, as exemplified by TD-MPC2. Specifically, the total world model trains an encoder \( z = h(s) \), a latent dynamics model \( z' = f(z, a) \), a reward predictor \( R(z, a) \), and a value function \( Q(z, a) \), where \( z \) denotes the latent state. Throughout the learning process, all states \( s \) are first encoded into latent representations \( z \). For notational simplicity, we continue to denote the inputs to the policy and value functions as \( s \), although they implicitly refer to the encoded latent states. The components \( h \), \( f \), \( R \), and \( Q \) are jointly trained using the TD loss:
\begin{equation}
\label{eq_model_loss}
\begin{split}
    \mathcal{L}_\text{model} = \mathbb{E}_{(s, a, r, s')_{0:H}} \left[ \sum_{t=0}^H \gamma^t \left( \left\| f(z_t, a_t) - \text{sg}(h(s_t')) \right\|_2^2 + \text{CE}(R_t, r_t) + \text{CE}(Q_t, q_t) \right) \right],
\end{split}
\end{equation}
where $q_t$ is the target value computed using the policy network $\pi$, $\text{sg}$ denotes the stop-gradient operator and CE denotes the cross entropy loss function. Then following standard off-policy RL, the policy $\pi$ is updated by maximizing predicted Q-values, i.e., $\mathcal{L}_{\text{policy}} = - \mathbb{E}_{(s, a, r, s')_{0:H}} \sum_{t=0}^{H-1} Q(s, \pi(s))$.

Despite being trained purely on off-policy data, this world model empowers the planner, such as MPPI, to perform effective predictive optimization. We refer to the resulting behavior policy executed at each timestep as $\beta$, which can be understood as $\pi + $MPPI with world model.
By seamlessly integrating model-based planning into the off-policy learning pipeline, this approach yields high-quality training interactions and enhances sample efficiency. 

However, this paradigm inevitably suffers from a phenomenon known as \textit{actor divergence}. During data collection, the agent interacts with the environment using a planner-augmented policy $\beta = \pi + \text{MPPI}$, which refines policy actions using model rollouts. This behavior policy $\beta$ may exhibit highly deterministic or multi-modal behavior, resulting in a state-action distribution $d^\beta(s, a)$ that may significantly diverges from that of the network policy $\pi$. 
This mismatch breaks the assumptions of standard off-policy learning and can lead to instability in training process. Specifically, 
such an actor divergence leads to following two major challenges:

\textbf{(1) Distribution shift in value learning.}  
The value function is trained to minimize the Bellman error over samples drawn from the behavior distribution:
$
\mathbb{E}_{(s, a) \sim d^\beta} \left[ Q(s, a) - \mathcal{T}^\pi Q(s, a) \right]^2,
$
where $\mathcal{T}^\pi Q(s, a) = r(s, a) + \gamma \mathbb{E}_{a' \sim \pi(s')}[Q(s', a')]$ denotes the Bellman backup under policy $\pi$.  
However, when $d^\beta$ assigns little probability mass to regions where $\pi$ places significant density, the value function is only optimized on a narrow subset of the state-action space. This mismatch can result in biased and generally overconfident estimates in out-of-distribution regions, undermining the accuracy and reliability of value learning.

\textbf{(2) Unreliable policy improvement.}  
The policy $\pi$ is typically optimized by maximizing the expected value under $Q$.  
However, due to the distribution shift mentioned above, $Q$ may be inaccurate, especially in regions where $\pi$ assigns high probability but are poorly represented in the behavior distribution $\beta$. 
This bias will result in unstable policy updates and poor performance.

In summary, \textit{actor divergence} inevitably occurs when online planning meets off-policy RL, which undermines the core assumption of distributional consistency in off-policy learning paradigm and degrades both model accuracy and policy improvement.

\begin{figure}[t]
  \centering
    \includegraphics[width=1.0\textwidth,trim={0.28cm 0.0cm 0.28cm 0.0cm},   clip]{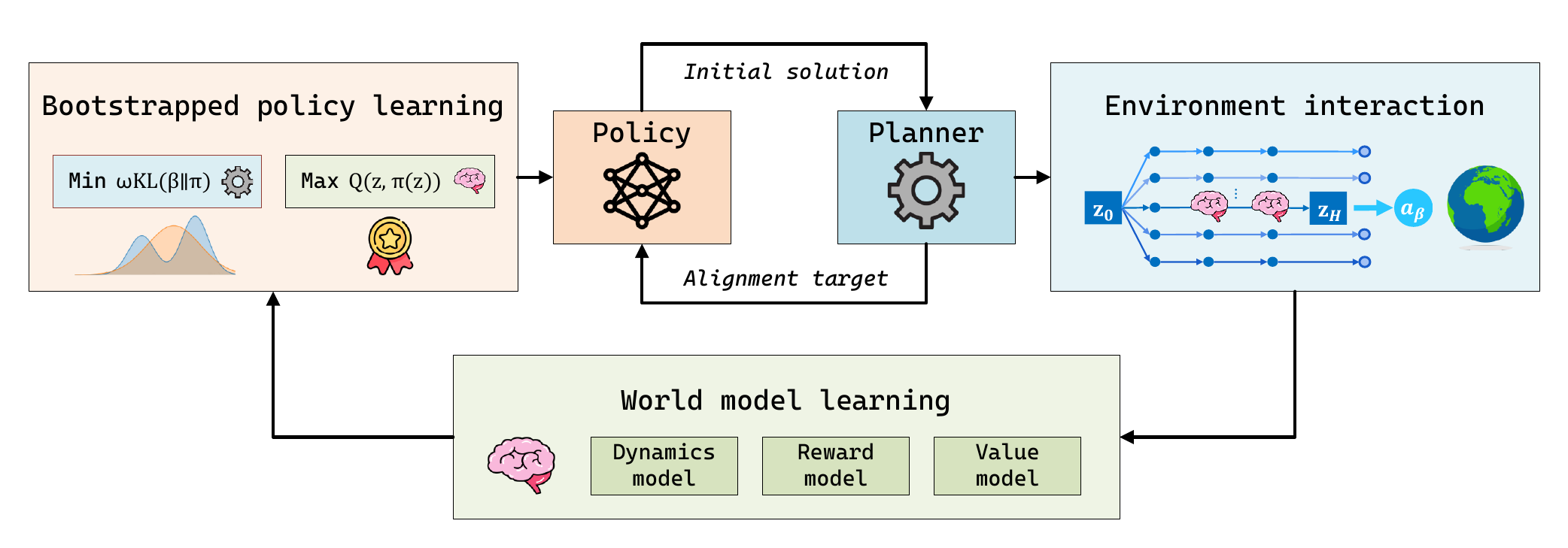}
  \caption{\textbf{Overview of the proposed BOOM algorithm.} BOOM consists of three key components: a policy, a planner, and a world model. The policy and planner form a bootstrap loop where the policy provides the planner with an initial solution, and the planner in turn guides the policy via alignment. The world model plays a dual role: it enables the planner to perform receding horizon control for collecting  high-quality trajectories, and it allows the policy to utilize Q-values for effective performance improvement.}
  \label{fig_method}
\end{figure}

\subsection{Bootstrap Off-policy with World Model}

To mitigate actor divergence caused by collecting data with a planning-augmented behavior policy (e.g., $\beta = \pi + \text{MPPI}$), we propose a simple yet effective framework: \textbf{BOOM} (Bootstrap Off-policy with World Model) as shown in Figure \ref{fig_method}. It comprises three tightly coupled components—a policy, a planner, and a world model.
At the heart of BOOM is a bootstrap loop: the policy provides an initialization for the planner, while the planner refines this initialization through model predictive optimization and in turn bootstraps the policy via behavior alignment. The world model, trained in the TD-MPC2 style, serves a dual purpose: it enables the planner to simulate future trajectories for better control, and it supports the policy with value estimates for improvement.

The core of BOOM is the \emph{Bootstrap Alignment} objective—a likelihood-free regularization term that encourages the policy $\pi$ to align with planner-generated actions without requiring the likelihoods of planner's non-parametric distribution. This objective is further enhanced by a soft value-weighted mechanism that prioritizes high-return behaviors. We describe these two components in detail below.

\textbf{Likelihood-free alignment metric.}  
Since all actions stored in the replay buffer are generated by the planner policy $\beta$, we can directly imitate them to align the planner behaviors without impairing training efficiency. However, $\beta$ is a non-parametric sample-based planner whose action likelihood is intractable. This makes typical imitation learning metric, such as reverse KL divergence, theoretically inapplicable, as they rely on knowing $\beta(a \mid s)$.
To avoid this, we adopt a \emph{likelihood-free} approach by minimizing the forward KL divergence:
\begin{equation}
\mathrm{KL}(\beta \,\|\, \pi) = \mathbb{E}_{a \sim \beta(\cdot \mid s)} \left[ \log \frac{\beta(a \mid s)}{\pi(a \mid s)} \right] = \mathbb{E}_{a \sim \beta(\cdot \mid s)} \left[ \log \beta(a \mid s) \right] - \mathbb{E}_{a \sim \beta(\cdot \mid s)} \left[ \log \pi(a \mid s) \right].
\end{equation}
The first term depends only on $\beta$ and is constant with respect to the parameters of $\pi$. Therefore, we discard it during optimization, resulting in the following simplified loss:
\begin{equation}
\mathcal{L}_{\text{align}} = \mathbb{E}_{(s, a, r, s')_{0:H}} \left[ -\log \pi(a \mid s) \right],
\end{equation}
which encourages $\pi$ to assign higher probability to actions chosen by the planner, without requiring any access to $\beta(a \mid s)$. This likelihood-free formulation provides a simple, principled mechanism to distill the strengths of an online planner—yielding non-parametric action distributions—into a parametric policy.

\textbf{Soft Q-weighted mechanism.}  
To further enhance policy learning, we introduce a value-guided alignment objective that prioritizes high-return behaviors, drawing inspiration from the planner’s action selection principle—where candidate actions are first sampled and then selected based on their value-weighted probabilities. Given a batch of off-policy transitions $\{(s_i, a_i)\}_{i=1}^N$ replayed from buffer, we define a soft target distribution over actions using the current Q-function:
$
p \propto \exp\left({Q}/{\tau}\right),
$
where $\tau > 0$ controls the distribution sharpness and is set to $1$ by default. Normalization across the batch yields weights
$
w_i = {\exp(Q_i/\tau)}/{\sum_{j=1}^N \exp(Q_i/\tau)}.
$
Now the soft Q-weighted alignment loss becomes
\begin{equation}
    \mathcal{L}_{\text{align}} = \mathbb{E}_{(s, a, r, s')_{0:H}} \sum_{t=0}^{H-1} \sum_{i=1}^N w_i \left[-\log \pi(a_i \mid s_i) \right].
\end{equation}
This mechanism ensures that the policy assigns higher probability to high-value actions, effectively guiding it toward promising regions. Besides, since these actions stored in the replay buffer may vary in quality, this value-weighting ensures that the policy prioritizes the most beneficial experiences, thus accelerating learning while accommodating the variability in the planner’s past actions. The Q function here can also be replaced by other critics, such as state value V or advantage A, here we think Q is the most convenient one to access.


\textbf{Bootstrapped policy objective.}  
By combining the above two techniques, we integrate the alignment term into the standard policy loss to obtain the final bootstrapped policy objective:
\begin{equation}
    \mathcal{L}_{\text{policy}} =  - \mathbb{E}_{(s, a, r, s')_{0:H}} \sum_{t=0}^{H-1}  \bigg[ Q(s, \pi(s)) + \lambda_{\text{align}} \cdot \mathcal{L}_{\text{align}} \bigg],
    \label{eq_policy_loss}
\end{equation}
where $\lambda_{\text{align}}$ is a tunable coefficient. This bootstrapped objective encourages the policy not only to improve with respect to its own value estimates ($\max~Q$) but also to stay aligned with the high-quality planner actions found in the replay buffer ($\min~\mathcal{L}_\text{q-align}$). The complete pseudocode of our BOOM is presented in Algorithm \ref{alg:boom}.  

\paragraph{Discussion on why the world model learning is improved.}
The reasons are two-fold. \textit{(1) Improved value learning.}
By aligning the policy with the planner, we reduce the discrepancy between the collected data and the actual policy behavior. This improves the distributional matching during training, enabling the value estimator to learn from more consistent and policy-relevant trajectories. As a result, the predicted values used in planning become more accurate and reliable.
\textit{(2) Improved representation, reward, and dynamics learning.}
We adopt a TD-style learning objective that jointly optimizes the value function along with the other components of the world model as shown in \eqref{eq_model_loss}.
As the value predictions become more accurate due to better distributional match, the gradients flowing into the encoder, dynamics model, and reward predictor become more informative, leading to improved overall model quality.



\begin{algorithm}[t]
\caption{\textbf{BOOM}: Bootstrap Off-policy with World Model}
\label{alg:boom}
\textbf{Input:} Policy $\pi_\theta$, encoder $h_\xi$, dynamics $f_\psi$, reward $R_\omega$, value $Q_{\phi}$, planner $\mathcal{P}$ \\
\textbf{Initialize:} $\pi_\theta$, $h_\xi$, $f_\psi$, $R_\omega$, $Q_{\phi}$
\begin{algorithmic}[1]
\vspace{5pt}
    \STATE \textbf{\texttt{// Warmup (World Model Pretraining)}}
    \STATE Interact with environment using random actions: $(r, s', \texttt{done}) \leftarrow \texttt{env.step}(a_\text{rand})$
    \STATE Store random transitions $(s, a_\text{rand}, r, s')$ into replay buffer $\mathcal{D}$
    \STATE  Update $h_\xi$, $f_\psi$, $R_\omega$, $Q_{\phi}$ by minimizing model loss $\mathcal{L}_{\text{model}}$ in \eqref{eq_model_loss}
    \vspace{5pt}
    \FOR{each iteration}
        \STATE \texttt{// Data Collection (Using Planner)}
        \STATE Encode current observation: $z = h_\xi(s)$
        \STATE Plan: $a_\beta \sim \beta = \mathcal{P}(\pi_\theta, f_\psi, R_\omega, Q_\phi, z)$
        \STATE Interact with environment: $(r, s', \texttt{done}) \leftarrow \texttt{env.step}(a_\beta)$
        \STATE Store transition $(s, a_\beta, r, s')$ into replay buffer $\mathcal{D}$

        \vspace{5pt}
        \STATE \texttt{// World Model and Policy Learning}
        \STATE Replay rollout batch $\{(s_t, a_\beta, r_t, s_{t+1})_{t=0}^{H-1}\} \sim \mathcal{D}$

        \STATE Update $h_\xi$, $f_\psi$, $R_\omega$, $Q_{\phi}$ by minimizing model loss $\mathcal{L}_{\text{model}}$ in \eqref{eq_model_loss}


        \STATE Update $\pi_\theta$ by minimizing bootstrapped policy loss $\mathcal{L}_\text{policy}$ in \eqref{eq_policy_loss}
    \ENDFOR
\end{algorithmic}
\end{algorithm}

\subsection{Theoretical Analysis}

We provide theoretical guarantees for bootstrap alignment in addressing \textit{actor divergence}—the mismatch between the planner collected data and the policy actual behavior. By minimizing the divergence $\mathrm{KL}(\beta \| \pi)$, we establish theoretical bounds on the return gap and Q-value deviation, ensuring stable and efficient policy learning.

\begin{theorem}[Bootstrap Alignment Controls Return Gap]
\label{theorem_return_gap}
Let $\beta(a \mid s)$ be the behavior policy, $\pi(a \mid s)$ be the learned policy, and $d^\beta(s)$ the state distribution induced by $\beta$. Assume the per-step reward satisfies $|r(s,a)| \le R_{\max}$ and the discount factor $\gamma \in [0,1)$. Then for any state $s$, if
$
\mathrm{KL} (\beta \| \pi ) \le \varepsilon,
$
the following return gap bound holds:
\begin{equation}
\bigl| J(\beta) - J(\pi) \bigr| \le \frac{R_{\max}}{1 - \gamma} \sqrt{2\varepsilon}.
\end{equation}
\end{theorem}
\begin{proof}
    See Appendix \ref{proof_theorem_1}
\end{proof}

The first theorem shows that bootstrap alignment ensures a small return gap between $\beta$ and $\pi$, meaning staying close to the planner avoids performance drops from distribution mismatch.

\begin{theorem}[Bootstrap Alignment Controls Q-Value Gap]
\label{theorem_q_gap}
Assume that for any state $s$, the learned policy $\pi(a \mid s)$ and the planner $\beta(a \mid s)$ satisfy $\mathrm{KL}(\beta \| \pi) \le \varepsilon$, and that $Q(s,a)$ is $L_Q$-Lipschitz continuous in $a$. Then for any state $s$, the expected Q-value difference is bounded as:
\begin{equation}
    \left| Q(s, a_\beta) - Q(s, a_\pi) \right| 
    \le L_Q \cdot \| a_\beta - a_\pi \|_2 \le L_Q \cdot D(\varepsilon),
\end{equation}
where $D(\varepsilon)$ is an upper bound on the 2-norm distance $\| a_\beta - a_\pi \|_2$ with $a_\beta \sim \beta(s)$ and $a_\pi \sim \pi(s)$.
In general, the action distribution of MPPI can be approximated by a Gaussian Mixture Model (GMM) as $\beta(s) = \sum_{i=1}^K w_i \mathcal{N}(\mu_i, \Sigma_i)$ with weights $w_i \ge 0$, $\sum_{i=1}^K w_i = 1$, and $\pi(s) = \mathcal{N}(\mu_\pi, \Sigma_\pi)$ is a Gaussian policy. Denote the maximum eigenvalue of each $\Sigma_i$ as $\Lambda_i := \Lambda(\Sigma_i)$ and that of $\Sigma_\pi$ as $\Lambda(\Sigma_\pi)$. Then for any $\delta \in (0,1)$, with probability at least $1 - \delta$ over $a_\beta \sim \beta(s)$ and $a_\pi \sim \pi(s)$, the deviation bound satisfies:
\begin{equation}
    D(\varepsilon) \le \min \left( 2 \sqrt{d}, \
\max_i \Big( \sqrt{2 \Lambda_i \log \tfrac{K}{\delta}} + \sqrt{2 \varepsilon \Lambda(\Sigma_\pi) / w_i} \Big) + 
    \sqrt{2 \Lambda(\Sigma_\pi) \log \tfrac{1}{\delta}} 
\right).
\end{equation}

Here, $d$ is the dimensionality of the normalized action space.
\end{theorem}

\begin{proof}
    See Appendix \ref{proof_theorem_2}
\end{proof}

The second theorem bounds the Q-value difference under bootstrap alignment, showing that value overestimation—commonly seen when the policy strays into poorly covered regions—is effectively controlled by alignment. This avoids misleading policy updates and stabilizes training.

Together, these insights justify that bootstrap alignment mitigates \textit{actor divergence} by keeping the learned policy close to the behavior policy, preventing large discrepancies in value estimates. It retains the benefits of off-policy learning without requiring access to behavior policy likelihoods, making it compatible with modern planners and applicable in complex, high-dimensional settings.

\section{Experiments}
\label{sec_exp}
\begin{figure}[t]
  \centering

  \begin{subfigure}[b]{0.245\textwidth}
    \includegraphics[width=\textwidth,trim={0.28cm 0.25cm 0.28cm 0.25cm},   clip]{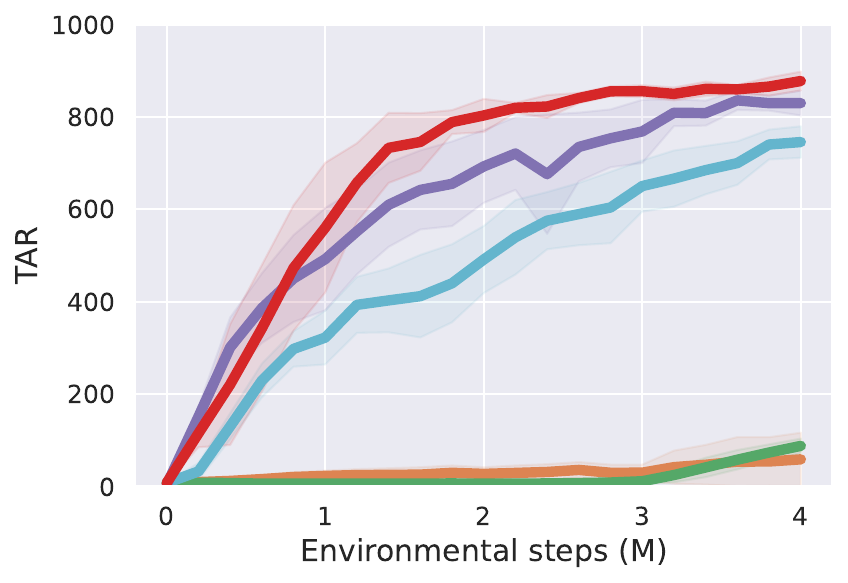}
    \caption{\textbf{AVG. DMC Suite}}
    \label{fig:dmc_total_average}
  \end{subfigure}
  \begin{subfigure}[b]{0.245\textwidth}
    \includegraphics[width=\textwidth,trim={0.28cm 0.25cm 0.28cm 0.25cm},   clip]{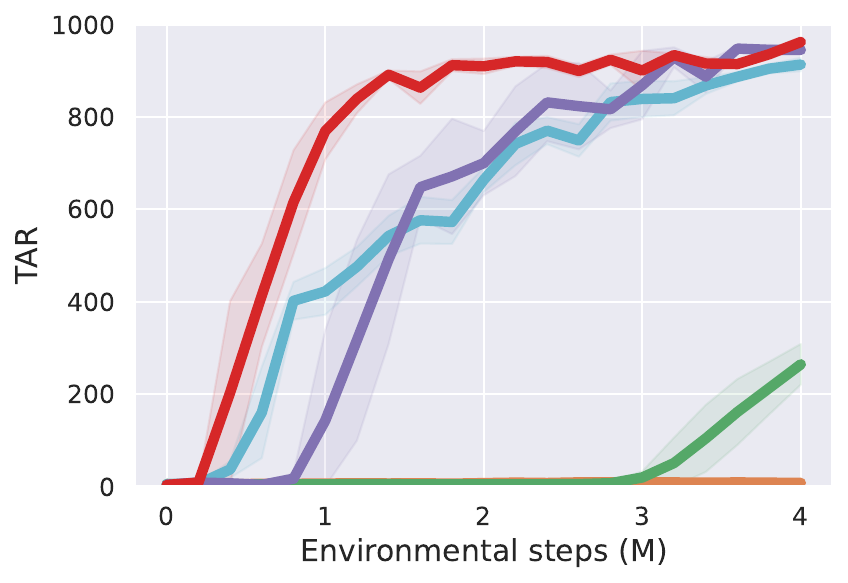}
    \caption{Humanoid-stand}
    \label{fig:humanoid-stand}
  \end{subfigure}
  \begin{subfigure}[b]{0.245\textwidth}
    \includegraphics[width=\textwidth,trim={0.28cm 0.25cm 0.28cm 0.25cm},   clip]{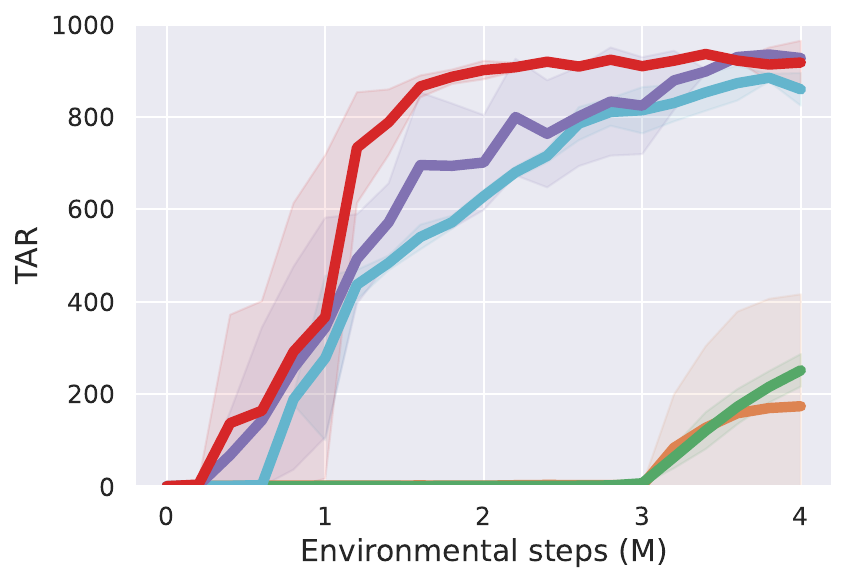}
    \caption{Humanoid-walk}
    \label{fig:humanoid-walk}
  \end{subfigure}
  \begin{subfigure}[b]{0.245\textwidth}
    \includegraphics[width=\textwidth,trim={0.28cm 0.25cm 0.28cm 0.25cm},   clip]{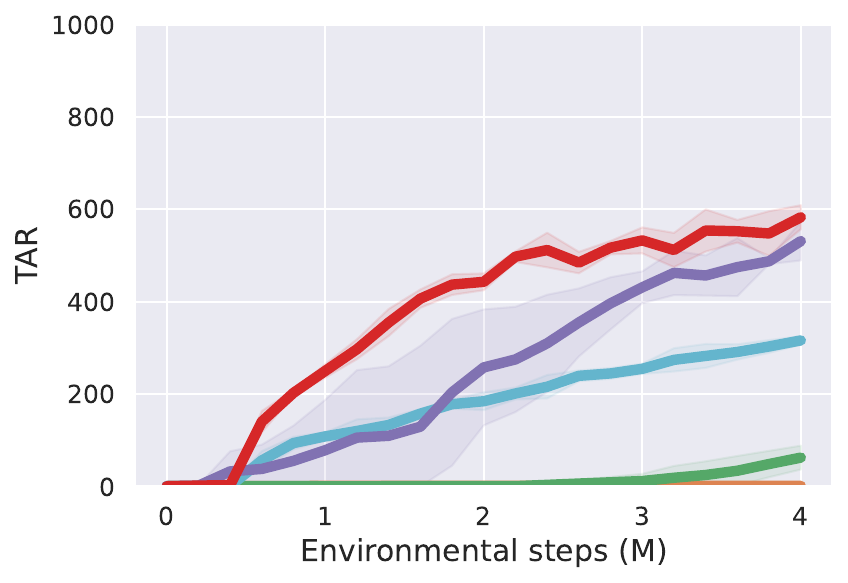}
    \caption{Humanoid-run}
    \label{fig:humanoid-run}
  \end{subfigure} \\
  \begin{subfigure}[b]{0.245\textwidth}
    \includegraphics[width=\textwidth,trim={0.28cm 0.25cm 0.28cm 0.25cm},   clip]{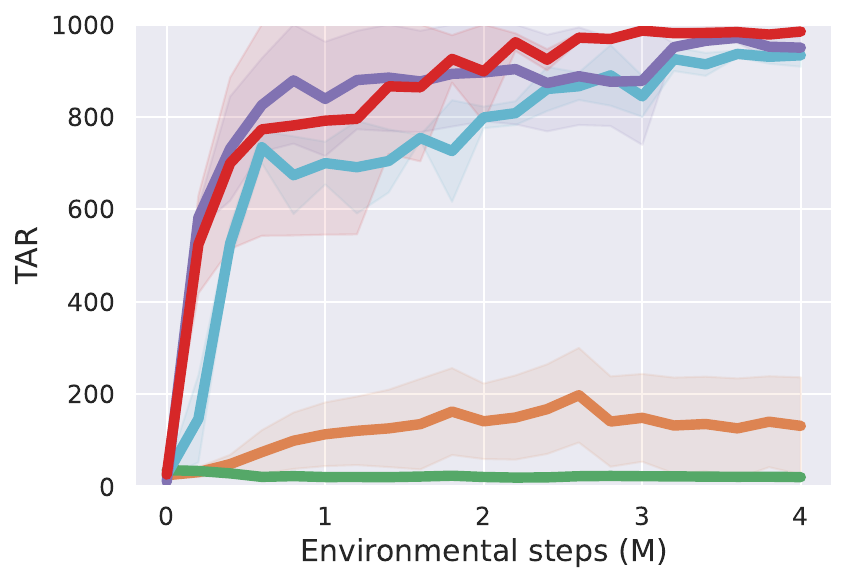}
    \caption{Dog-stand}
    \label{fig:dog-stand}
  \end{subfigure}
  \begin{subfigure}[b]{0.245\textwidth}
    \includegraphics[width=\textwidth,trim={0.28cm 0.25cm 0.28cm 0.25cm},   clip]{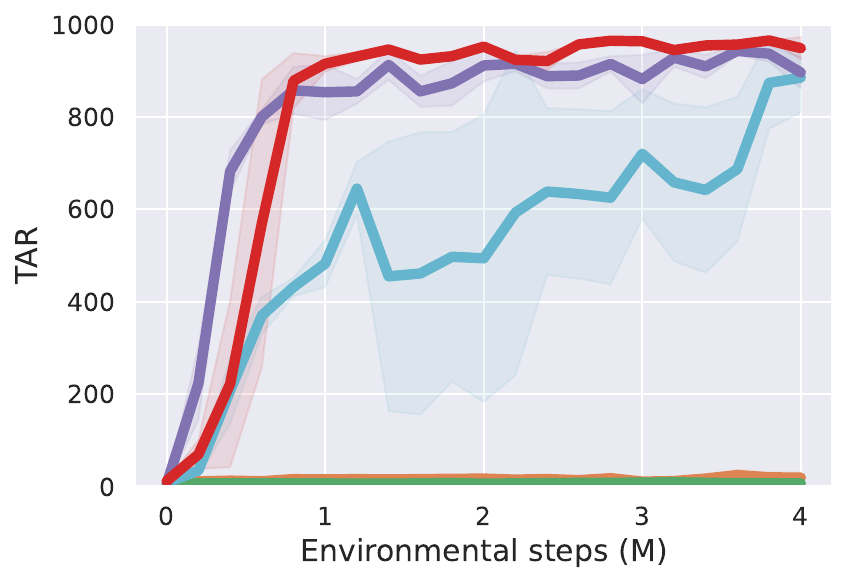}
    \caption{Dog-walk}
    \label{fig:dog-walk}
  \end{subfigure}
  \begin{subfigure}[b]{0.245\textwidth}
    \includegraphics[width=\textwidth,trim={0.28cm 0.25cm 0.28cm 0.25cm},   clip]{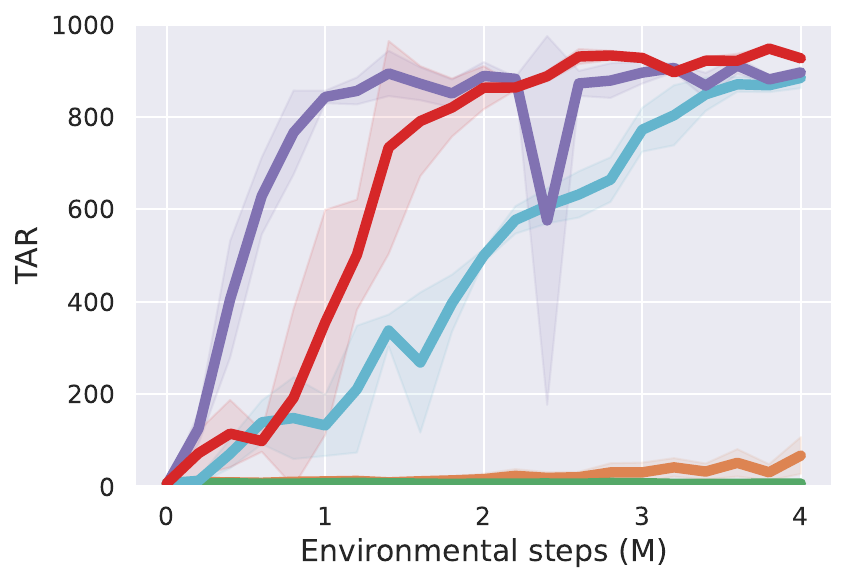}
    \caption{Dog-trot}
    \label{fig:dog-trot}
  \end{subfigure}
  \begin{subfigure}[b]{0.245\textwidth}
    \includegraphics[width=\textwidth,trim={0.28cm 0.25cm 0.28cm 0.25cm},   clip]{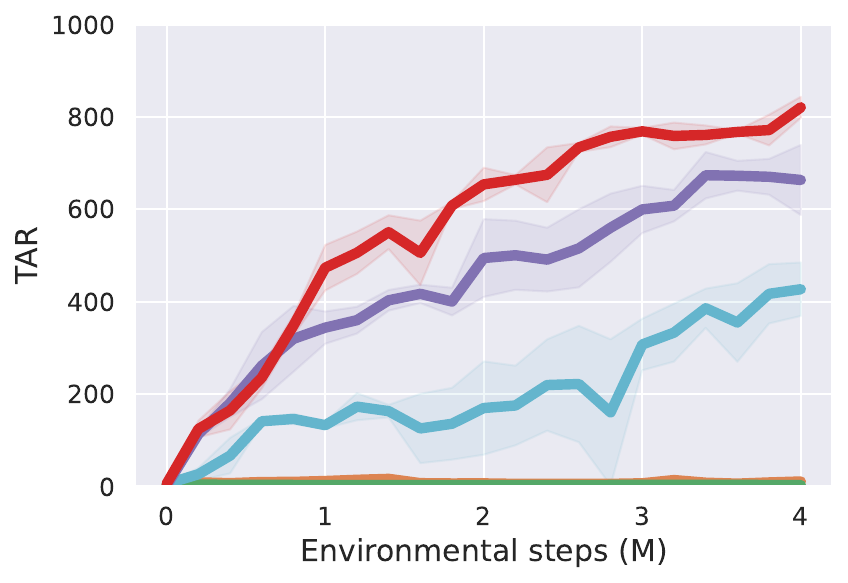}
    \caption{Dog-run}
    \label{fig:dog-run}
  \end{subfigure} \\
  \begin{subfigure}[b]{0.245\textwidth}
    \includegraphics[width=\textwidth,trim={0.28cm 0.25cm 0.28cm 0.25cm},   clip]{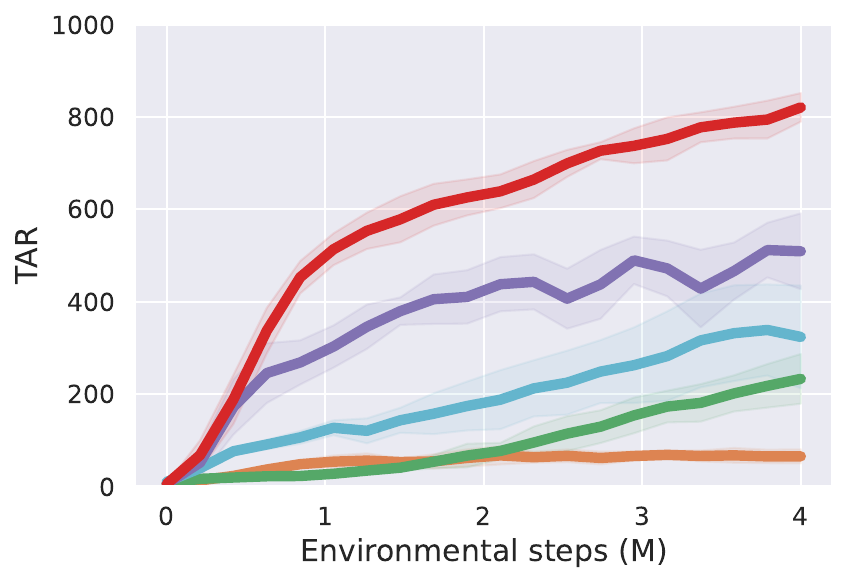}
    \caption{\textbf{AVG. Human. Bench}}
    \label{fig:H-Bench Average}
  \end{subfigure}
    \begin{subfigure}[b]{0.245\textwidth}
    \includegraphics[width=\textwidth,trim={0.28cm 0.25cm 0.28cm 0.25cm},   clip]{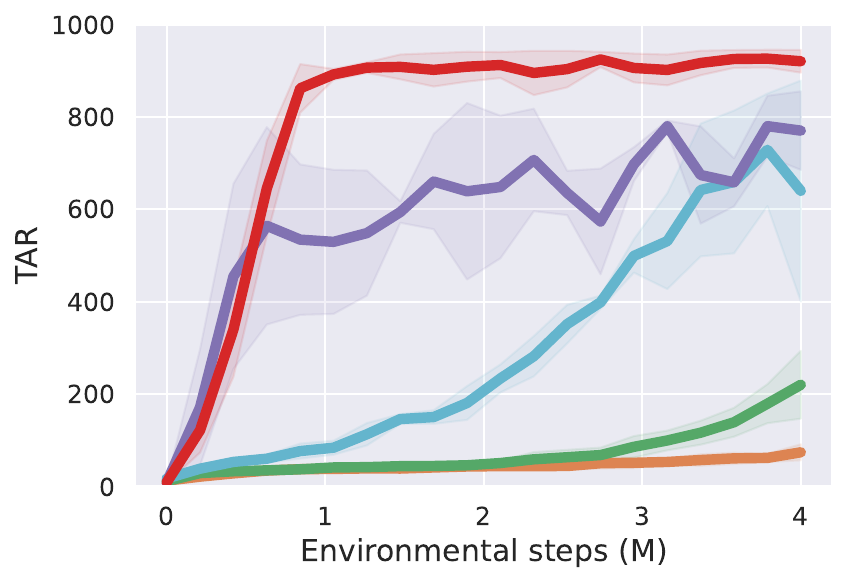}
    \caption{H1hand-stand}
    \label{fig:stand}
  \end{subfigure}
  \begin{subfigure}[b]{0.245\textwidth}
    \includegraphics[width=\textwidth,trim={0.28cm 0.25cm 0.28cm 0.25cm},   clip]{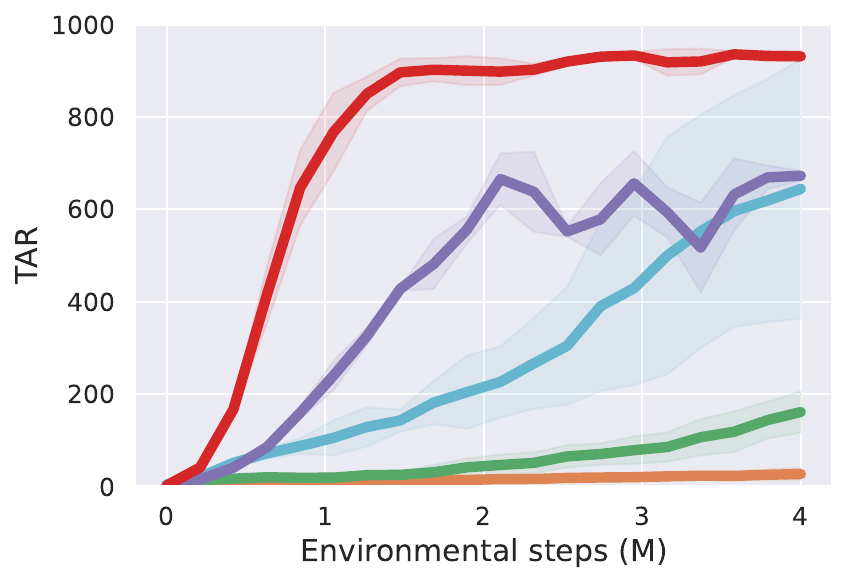}
    \caption{H1hand-walk}
    \label{fig:walk}
  \end{subfigure}
  \begin{subfigure}[b]{0.245\textwidth}
    \includegraphics[width=\textwidth,trim={0.28cm 0.25cm 0.28cm 0.25cm},   clip]{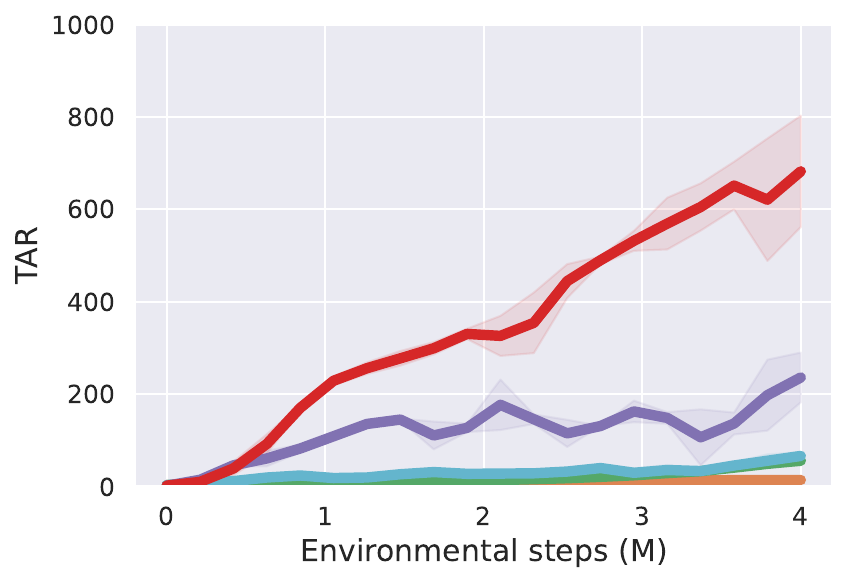}
    \caption{H1hand-run}
    \label{fig:run}
  \end{subfigure} \\
  \begin{subfigure}[b]{0.245\textwidth}
    \includegraphics[width=\textwidth,trim={0.28cm 0.25cm 0.28cm 0.25cm},   clip]{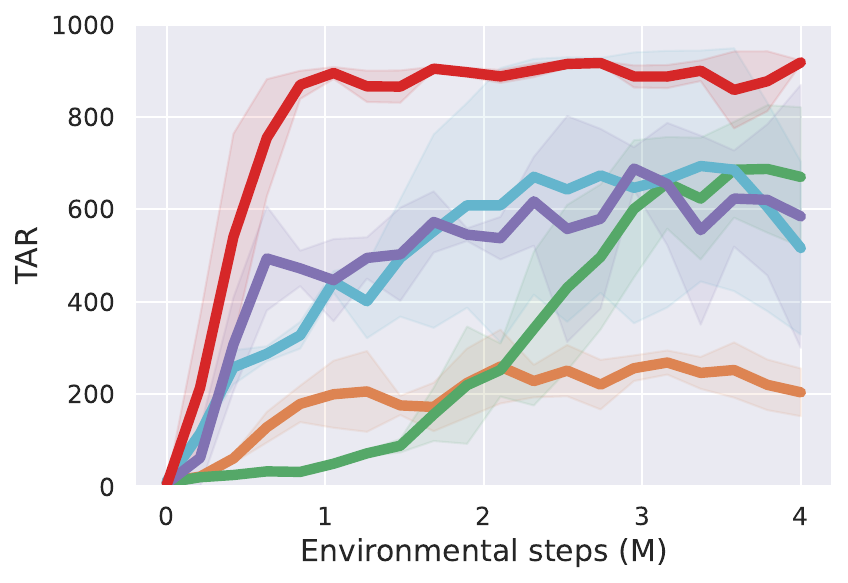}
    \caption{H1hand-sit}
    \label{fig:sit}
  \end{subfigure}
  \begin{subfigure}[b]{0.245\textwidth}
    \includegraphics[width=\textwidth,trim={0.28cm 0.25cm 0.28cm 0.25cm},   clip]{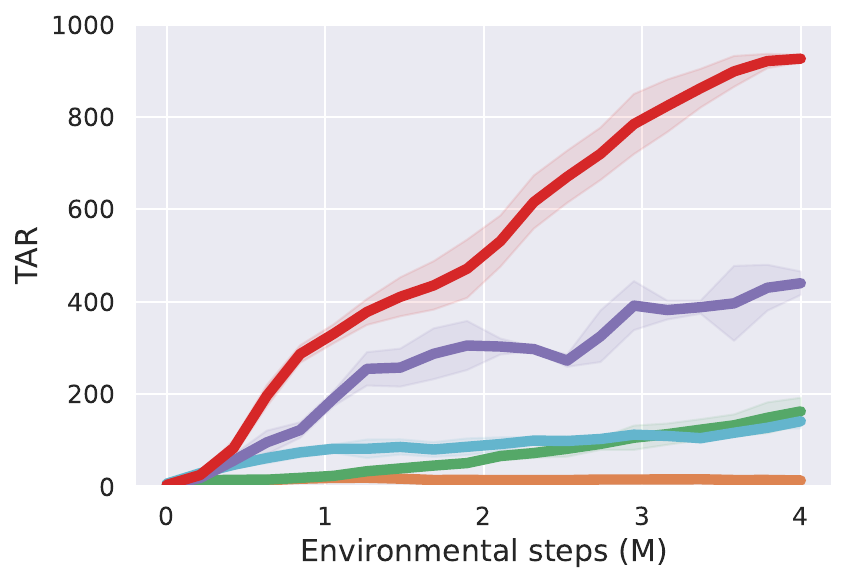}
    \caption{H1hand-slide}
    \label{fig:slide}
  \end{subfigure}
  \begin{subfigure}[b]{0.245\textwidth}
    \includegraphics[width=\textwidth,trim={0.28cm 0.25cm 0.28cm 0.25cm},   clip]{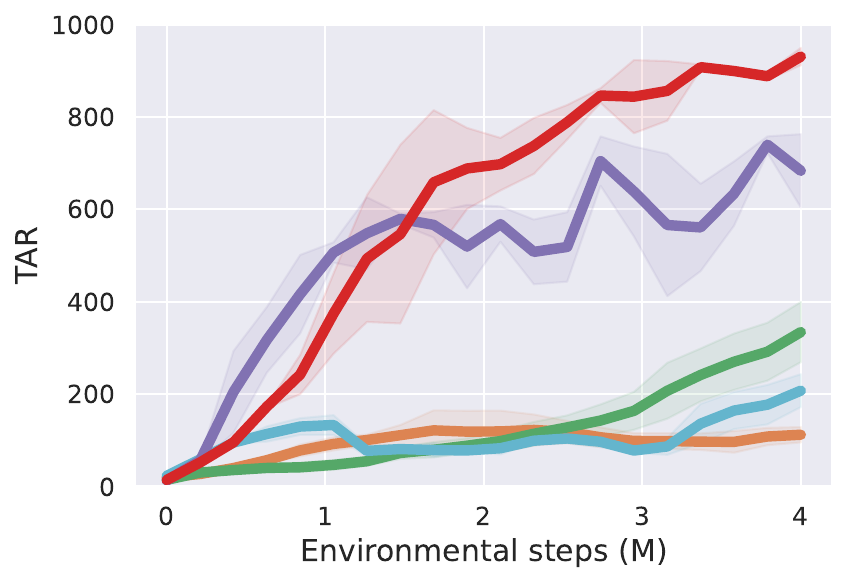}
    \caption{H1hand-pole}
    \label{fig:pole}
  \end{subfigure}
  \begin{subfigure}[b]{0.245\textwidth}
    \includegraphics[width=\textwidth,trim={0.28cm 0.25cm 0.28cm 0.25cm},   clip]{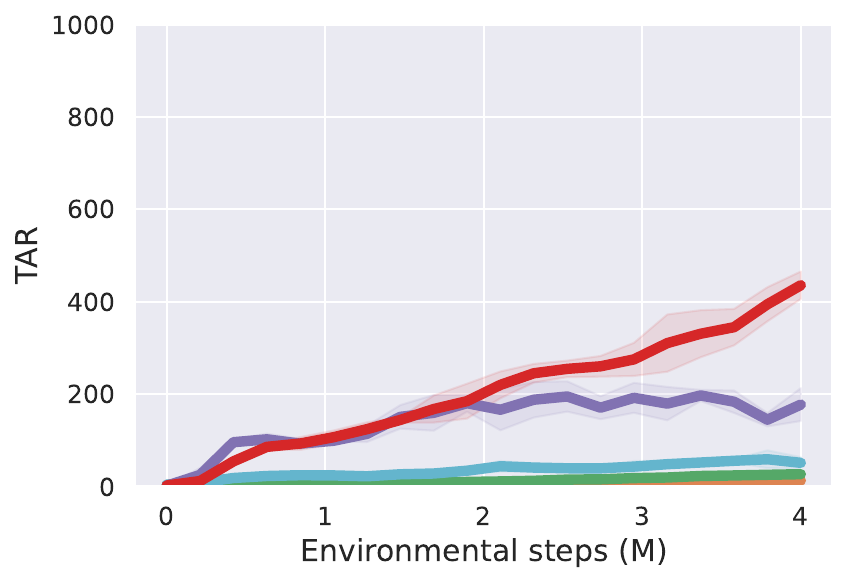}
    \caption{H1hand-hurdle}
    \label{fig:hurdle}
  \end{subfigure}
   \begin{subfigure}[b]{0.95\textwidth}
    \includegraphics[width=\textwidth,trim={0.0cm 0.cm 1.2cm 0.0cm},   clip]{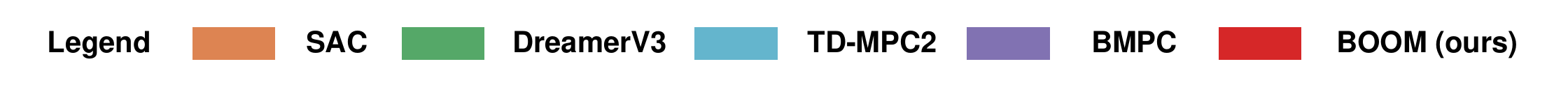}
  \end{subfigure}
  \caption{\textbf{Training curves on benchmarks.} The solid lines represent the mean, while the shaded
regions indicate the confidence interval over three runs. The average performance curves for the two benchmarks appear at the left corner of the 1st and 3rd rows, respectively, highlighted in \textbf{bold}.}
  \label{fig_training_curves}
\end{figure}

\subsection{Experimental Setup}

\paragraph{Baselines.} We choose four representative online RL algorithms as our baselines: 
(1) \textbf{SAC}~\cite{haarnoja2018soft}: the state-of-the-art model-free off-policy RL algorithm under the maximum entropy framework; 
(2) \textbf{DreamerV3}~\cite{hafner2025mastering}: the state-of-the-art imagination-driven model-based RL algorithm that learns from imaginary rollouts generated by a world model. To enable a comprehensive comparison, we evaluate DreamerV3 under two different interaction budgets: 2M and 10M iterations; 
(3) \textbf{TD-MPC2}~\cite{hansen2023td}: a planning-driven model-based RL algorithm that performs online planning via an MPPI planner; 
(4) \textbf{BMPC}~\cite{wang2025bootstrapped}: an improved variant of TD-MPC2, where the policy is trained solely to imitate the actions generated by the MPPI with an extra relabeling mechanism to update the historical actions in the replay buffer with the latest planner.

\paragraph{Benchmarks.} We evaluate our method on a challenging benchmark of 14 high-dimensional locomotion tasks drawn from the DeepMind Control Suite (DMC)~\cite{tassa2018deepmind} and the recently proposed Humanoid Bench (H-Bench)~\cite{sferrazza2024humanoidbench}. The chosen 7 DMC tasks feature two most complex agents—humanoid (67/21 state/action dims) and dog (223/38)—that demand sophisticated balance and coordination. The other 7 H-Bench tasks raise the difficulty further with long-horizon, goal-directed tasks on the Unitree H1hand robot (151/61), such as walking over  slides, traveling over a pole forest without collision, and continuously crossing hurdles.  Detailed descriptions are listed in Appendix~\ref{benchmark}.

\textbf{Implementation details.} The detailed hyperparameters and reproducibility statement of other baselines are documented in Appendix~\ref{hyperparameters}.

\subsection{Experimental Results}
All the training curves are shown in Figure \ref{fig_training_curves} and the detailed numerical results are listed in Table~\ref{tab_tar}. Our method, {BOOM}, consistently delivers the best Total Average Return (TAR) across all 14 high-dimensional locomotion tasks. These tasks pose significant challenges due to their large state and action spaces, yet BOOM demonstrates remarkable stability and effectiveness.

\paragraph{Results on the DMC Suite.} Our BOOM achieves an average TAR of \textbf{877.7}, outperforming the previous best BMPC (\underline{835.8}) by a substantial \textbf{+5.0\%}, and exceeding TD-MPC2 by an even larger margin of \textbf{+17.7\%}. Notably, BOOM achieves new best results on all humanoid and dog tasks. In \textit{Humanoid-run}, BOOM outperforms the second-best method by \textbf{+9.7\%}, and in \textit{Dog-run}, it leads by a staggering \textbf{+21.8\%}. 
While SAC and DreamerV3 often fail on these high-dimensional control tasks and achieve near-zero performance, planning-based methods like TD-MPC2 and BMPC perform better but still suffer from instability and limited final returns. In contrast, BOOM learns faster and achieves much stronger final performance across all tasks.

\paragraph{Results on the Humanoid Bench.} Our BOOM again dominates with an average TAR of \textbf{820.6}. This marks a dramatic \textbf{+47.7\%} improvement over DreamerV3 (10M), which is (\underline{555.6}), and an even more impressive \textbf{+60.5\%} gain over BMPC. BOOM sets new records on every single task in the Humanoid Bench. For example, in \textit{H1hand-slide}, BOOM improves over the second-best method by \textbf{+110.5\%}, in \textit{H1hand-pole}, by \textbf{+25.8\%}, and in \textit{H1hand-hurdle}, by \textbf{+121.0\%}, .

In summary, compared to SAC's difficulty in scaling to complex control, DreamerV3's limitations in sample efficiency, and BMPC/TD-MPC2's occasional instability, our BOOM demonstrates clear and consistent advantages across the board.

\begin{table}[h]
\fontsize{8pt}{10pt}\selectfont
\setlength{\tabcolsep}{4.2pt} 
\centering
\caption{Total Average Return (TAR) on 7 DMC Suite tasks and 7 Humanoid Benchmark (H-Bench) tasks. Mean ± Std over 3 seeds. \textbf{Bold} = best, \underline{underlined} = second-best. Higher is better.
}
\label{tab_tar}
\begin{tabular}{lllllll}
\toprule
\textbf{Task} & {\textbf{SAC}} & \multicolumn{2}{c}{\textbf{DreamerV3 ( 2M \& 10M iters)}}  & \textbf{TD-MPC2} & \textbf{BMPC}  & \textbf{BOOM (ours)} \\
\midrule
Humanoid-stand & 9.0 $\pm$ 0.7    & 264.5 $\pm$ 44.1   & 717.0 $\pm$ 21.2  & 913.3 $\pm$ 14.7 & \underline{947.9} $\pm$ 4.4 & \textbf{962.1} $\pm$ 10.7 \\
Humanoid-walk  & 173.8 $\pm$ 242.2    & 251.5 $\pm$ 35.2   & 755.6 $\pm$ 25.2  & 884.8 $\pm$ 8.3  & \underline{935.1} $\pm$ 4.0 & \textbf{936.1} $\pm$ 3.3 \\
Humanoid-run   & 1.6 $\pm$ 0.1    & 62.5 $\pm$ 25.8   & 353.5 $\pm$ 33.2  & 316.2 $\pm$ 9.2  & \underline{531.2} $\pm$ 42.0 & \textbf{582.8} $\pm$ 26.0 \\
Dog-stand      & 197.6 $\pm$ 102.4 & 35.4 $\pm$ 10.8 & 35.4 $\pm$ 10.8   & 936.4 $\pm$ 7.6 & \underline{971.3} $\pm$ 11.0 & \textbf{986.8} $\pm$ 1.8 \\
Dog-walk       & 24.7 $\pm$ 11.3   & 9.1 $\pm$ 0.6   & 9.1 $\pm$ 0.6     & 885.0 $\pm$ 74.8 & \underline{942.9} $\pm$ 9.6  & \textbf{965.4} $\pm$ 0.3 \\
Dog-trot       & 67.1 $\pm$ 39.9  & 7.9 $\pm$ 0.7   & 8.4 $\pm$ 0.9     & 884.4 $\pm$ 22.2 & \underline{911.3} $\pm$ 18.2 & \textbf{947.9} $\pm$ 4.7 \\
Dog-run        & 16.5 $\pm$ 8.5  & 4.3 $\pm$ 3.2   & 4.3 $\pm$ 3.2     & 427.0 $\pm$ 57.9 & \underline{673.7} $\pm$ 50.2 & \textbf{820.7} $\pm$ 23.0 \\
\midrule
\textbf{AVG. DMC Suite}  
& 58.8 $\pm$ 57.7 & 87.9 $\pm$ 15.8 & 269.0 $\pm$ 13.6 & 745.6 $\pm$ 34.1 & \underline{835.8} $\pm$ 20.7 & \textbf{877.7} $\pm$ 20.3 \\
\midrule
H1hand-stand       & 74.1 $\pm$ 17.5   & 220.3 $\pm$ 73.5    & \underline{845.4} ± 27.3  & 728.7 $\pm$ 121.9  & 780.0 $\pm$ 65.8   & \textbf{926.1} $\pm$ 19.2 \\
H1hand-walk        & 27.0 $\pm$ 13.8   & 161.3 $\pm$ 44.5   & \underline{744.0} ± 28.7  & 644.2 $\pm$ 281.1  & 672.6 $\pm$ 10.4   & \textbf{935.4} $\pm$ 7.3 \\
H1hand-run         & 14.1 $\pm$ 1.4    & 55.8 $\pm$ 10.5    & \underline{622.4} ± 66.7  & 66.1 $\pm$ 8.1     & 236.0 $\pm$ 53.9   & \textbf{682.2} $\pm$ 120.6 \\
H1hand-sit         & 268.4 $\pm$ 26.1 & 687.3 $\pm$ 138.0 & \underline{699.1} ± 177.2 & 693.7 $\pm$ 249.9 & 688.2 $\pm$ 46.3   & \textbf{918.1} $\pm$ 4.2 \\
H1hand-slide       & 19.0 $\pm$ 5.9   & 162.6 $\pm$ 29.5    & 367.6 ± 29.7  & 141.3 $\pm$ 15.6   & \underline{440.1} $\pm$ 25.4   & \textbf{926.1} $\pm$ 8.0 \\
H1hand-pole        & 122.5 $\pm$ 33.5 & 334.3 $\pm$ 65.1   & 577.4 ± 62.3  & 207.5 $\pm$ 35.6   & \underline{739.9} $\pm$ 18.0   & \textbf{930.5} $\pm$ 18.9 \\
H1hand-hurdle      & 12.9 $\pm$ 2.7    & 26.6 $\pm$ 3.0    & 135.7 ± 6.1   & 59.0 $\pm$ 19.3    & \underline{197.1} $\pm$ 12.1   & \textbf{435.6} $\pm$ 29.8 \\
\midrule
\textbf{AVG. H-Bench.} 
& 68.5 $\pm$ 9.2 & 233.0 $\pm$ 53.9 & \underline{555.6} ± 49.5 & 338.8 $\pm$ 98.6  & 511.7 $\pm$ 59.2  & \textbf{820.6} $\pm$ 31.0 \\
\bottomrule
\end{tabular}
\end{table}

\subsection{Ablation Study}

We conduct three ablation studies to assess the contribution of key components in our framework:

\begin{figure}[h]
  \centering
  \begin{subfigure}[b]{0.32\textwidth}
    \includegraphics[width=\textwidth,trim={0.28cm 0.25cm 0.28cm 0.25cm},   clip]{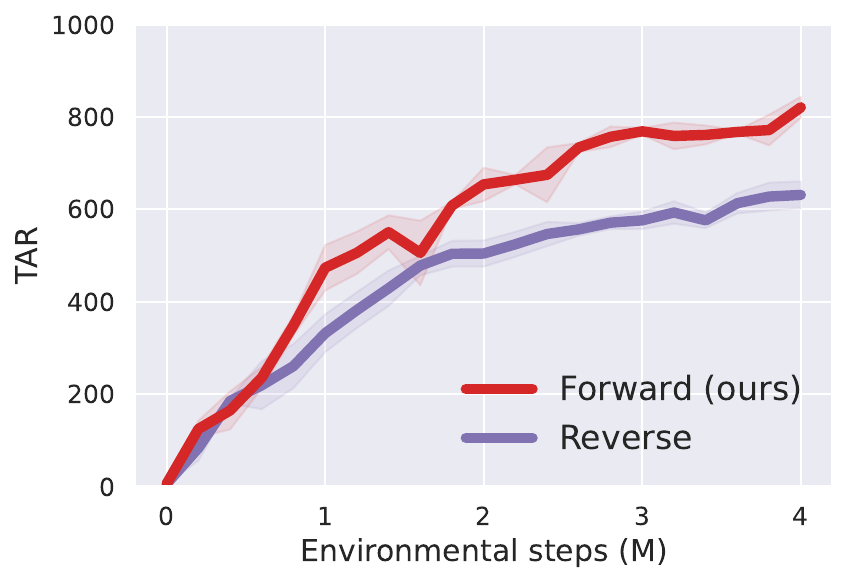}
    \captionsetup{width=0.75\linewidth} 
    \caption{Ablation study on the \\ \phantom{(b)}  alignment metric}
    \label{fig:Ablation2}
  \end{subfigure}
  \begin{subfigure}[b]{0.32\textwidth}
    \includegraphics[width=\textwidth,trim={0.28cm 0.25cm 0.28cm 0.25cm},   clip]{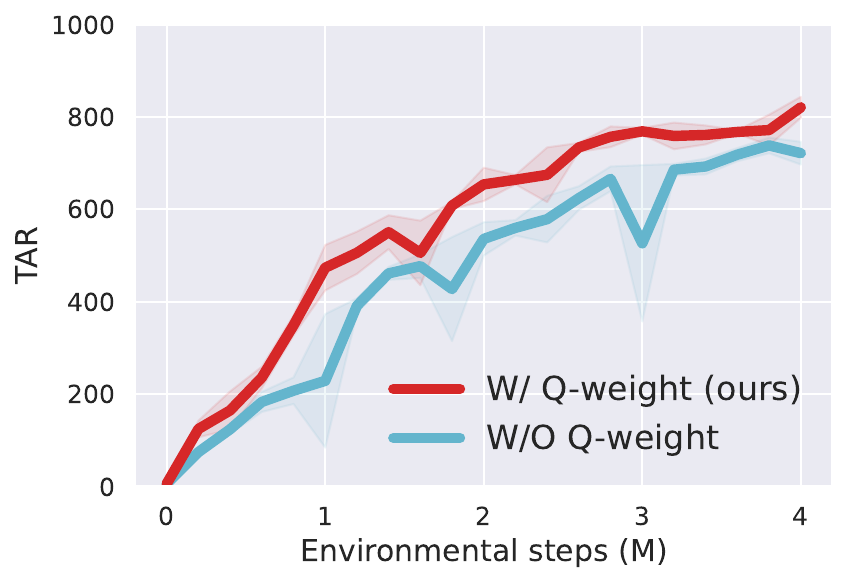}
    \captionsetup{width=0.75\linewidth} 
    \caption{Ablation study on the \\ \phantom{(a)} Q-weight mechanism}
    \label{fig:Ablation1}
  \end{subfigure}
  \begin{subfigure}[b]{0.32\textwidth}
    \includegraphics[width=\textwidth,trim={0.28cm 0.25cm 0.28cm 0.25cm},   clip]{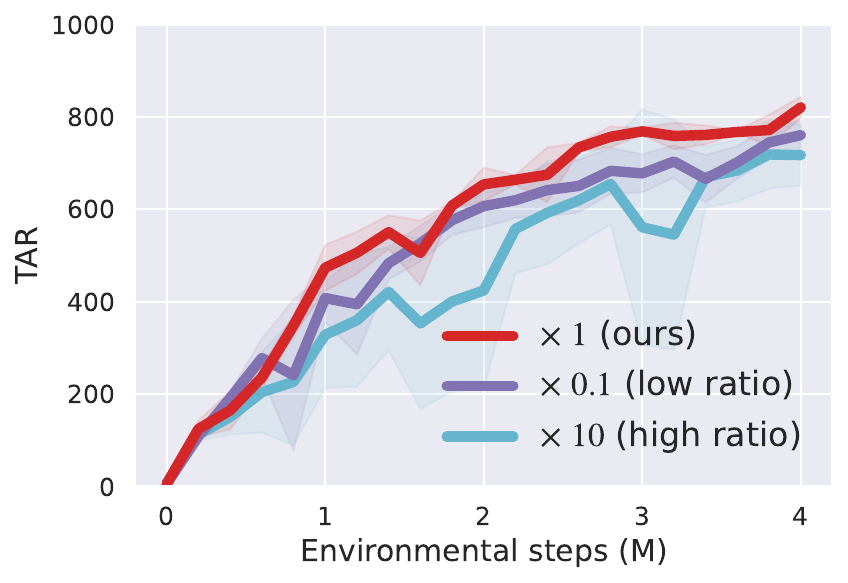}
    \captionsetup{width=0.75\linewidth} 
    \caption{Ablation study on the \\ \phantom{(c)} alignment coefficient}
    \label{fig:Ablation3}
  \end{subfigure}

  \caption{\textbf{Ablation study curves.} We select the \textit{Dog-run} task (223/38 state/action dims) in DMC Suite with the highest dimensionality to perform all ablation experiments.}
  \label{fig:example_3_images}
\end{figure}

\paragraph{Bootstrap alignment metric.}
We compare the common reverse KL divergence with our proposed likelihood-free forward KL. As shown in Figure~\ref{fig:Ablation2}, forward KL consistently shows higher returns, highlighting its strength in capturing the planner’s non-parametric distribution without requiring likelihood estimation.
Reverse KL, in contrast, relies on approximating the planner’s likelihood, which is not directly accessible. We estimate it using a Gaussian surrogate based on the value-weighted mean and variance of planner actions. The weaker performance also suggests that such approximations may be inaccurate and detrimental. Further discussion is provided in Appendix~\ref{compare_kl}.
\paragraph{Soft Q-weight mechanism.}
We replace the soft Q-weighting with uniform weighting to assess the benefit of leveraging the learned Q-function. As shown in Figure~\ref{fig:Ablation1}, incorporating Q-weights consistently accelerates training and improves final performance. This improvement stems from the ability of Q-weighting to handle the variability in action quality within the replay buffer by prioritizing high-value actions, thereby enabling more focused and efficient policy updates.
\paragraph{Alignment coefficient.}
We vary the alignment coefficient by testing $0.1\times$ and $10\times$ the default setting ($\lambda_\text{align} = \dim(\mathcal{A}) / 1000$ for DMC and $\dim(\mathcal{A}) / 50$ for Humanoid-Bench). {Results in Figure~\ref{fig:Ablation3}} show stable performance across this range, suggesting that BOOM is robust to this hyperparameter and does not require very sensitive tuning.

\section{Related Work}

Model-based RL, i.e., MBRL can be broadly categorized into planning-driven and imagination-driven approaches, depending on how the learned model is utilized during training.

\paragraph{Planning-driven MBRL.} 
Planning-driven methods use planner rather than policy itself to generate high-quality actions for environment interaction~\cite{curi2020efficient}.
Early work has shown that solely learning the value and then combining it with the planner can achieve good control performance ~\cite{nagabandi2018neural, kocijan2004gaussian, deisenroth2013gaussian, lowreyplan}. 
To further improve performance in high-dimensional tasks, recent approaches have combined online planning with policy learning, where the policy can provide a good initial solution to speed up planning\cite{janner2019trust, wang2019exploring, nguyen2021temporal, buckman2018sample, lin2025td}.  
LOOP~\cite{sikchi2022learning} takes SAC~\cite{haarnoja2018soft} as the backbone, and employs a planner under the policy behavior constraint for collecting samples. TD-MPC family~\cite{hansen2022temporal,hansen2023td} jointly learns model and
value function through TD-learning, achieving strong performance through both algorithmic innovations and implementation advances. It successfully delivers substantial gains over model-free baselines, particularly on complex benchmarks.  However, these methods inevitably encounter \textit{actor divergence}---a mismatch between the planner and the policy. Our approach, \textbf{BOOM}, addresses this challenge by tightly coupling planning and off-policy learning through a bootstrap loop that aligns the policy with the planner’s non-parametric action distribution via a Q-weighted likelihood-free alignment loss, preserving distributional consistency. A recent method BMPC~\cite{wang2025bootstrapped} simplifies the pipeline by discarding explicit policy optimization and directly imitating the planner to avoid \textit{actor divergence}. However, BMPC ignores the Q-function during training, resulting in lower policy learning efficiency and sensitivity to the variability of historical planner actions in the buffer. This leads to unstable learning, as reflected by its oscillatory training curves in our experiments. We acknowledge a concurrent and close work, TDM(PC)$^2$~\cite{lin2025td}, which similarly found that aligning the policy with the planner is beneficial. The major distinction lies in the design of the alignment objective: their approach follows a TD3+BC style using reverse KL, whereas ours is closer to AWAC using critic-guided weights and forward KL.

\paragraph{Imagination-driven MBRL.}  
Imagination-driven methods leverage a learned model to generate synthetic rollouts for policy and value updates~\cite{sutton1991dyna}. Modern approaches such as SimPLe~\cite{kaiser2019model}, IRIS~\cite{micheli2022transformers}, IDM~\cite{mu2021model}, and the Dreamer family~\cite{hafner2019dream,hafner2020mastering,hafner2023mastering} train latent dynamics models to support actor-critic learning entirely in imagination. These methods offer fast test-time execution and relatively high sample efficiency, but their performance is often limited by compounding model errors over long imagined rollouts. Among them, DreamerV3~\cite{hafner2025mastering} stands out as a leading representative, demonstrating strong performance across a range of tasks.
Unlike planning-based MBRL methods, DreamerV3 interacts by directly sampling actions from its learned policy, typically requiring more iterations to converge. For comprehensive comparison, we evaluate DreamerV3 under 2M and 10M iterations. While it improves with more interaction budget and outperforms TD-MPC2 and BMPC on certain tasks, our BOOM consistently outperforms all baselines across all tasks.

\paragraph{Compared to Offline RL.}
One might notice the similarity between BOOM and offline RL; however, the fundamental paradigms and practical implications of these two settings differ significantly. \textit{(1) Learning paradigm.} Offline RL centers on a fixed dataset, relying heavily on behavior cloning (BC) to restrict the policy within the dataset’s support~\cite{fujimoto2019off}. To cautiously improve policy performance, offline RL relaxes BC for poor actions, but always aims to keep the policy close to known data to avoid extrapolation error~\cite{prudencio2023survey, ran2023policy}.
BOOM centers on the policy itself as the optimization target. It seeks to maximize Q-values and align with planner-generated actions, both aimed at improving policy performance. Ultimately, the policy and planner co-adapt and converge to the optimal solution through online interaction.
\textit{(2) Learning objective.}
In offline RL, maximizing Q-values and BC often conflict~\cite{fujimoto2021minimalist}: Q-values outside the dataset distribution cannot be accurately estimated, so maximizing Q risks pushing the policy toward unsupported actions; BC pulls the policy back toward known actions. This tension forces a conservative balance~\cite{wang2022diffusion}.
In BOOM, the two objectives are largely complementary: the planner generally produces higher-quality actions than the policy~\cite{zhan2025bicriteria}. Aligning policy with the planner improves performance and strengthens policy-planner consistency, which in turn leads to more accurate Q estimates. More accurate Q-values then enable better policy improvement and allow the planner to generate higher-quality actions. This positive feedback loop bootstraps policy and planner consistently toward faster convergence to the optimal solution. 


\section{Conclusion}

We introduce BOOM, a model-based RL method that enhances the integration of planning and off-policy learning through bootstrap alignment. By leveraging the non-parametric planner actions not only for environment interaction but also for bootstrapping policy behavior via a Q-weighted likelihood-free alignment loss, BOOM mitigates the inevitably actor divergence issue in planning-driven model-based RL methods, improving both training stability and final performance while maintaining high time-efficiency and flexibility of off-policy learning paradigm. Experiments on tens of high-dimensional locomotion tasks show that BOOM consistently outperforms existing planning-driven and imagination-driven baselines.
We believe BOOM establishes a strong foundation with ample room for further improvement, such as more adaptive integration of max-Q and bootstrap alignment objectives, or the adoption of more expressive policy classes beyond diagonal Gaussians like diffusion models to unlock even higher performance and broader applicability.

Our work underscores two critical directions for advancing planning-driven MBRL. First, we emphasize that the learned value function remains the fundamental bottleneck; reliable improvement in its estimation accuracy consistently translates to enhanced planning and superior final asymptotic performance. Second, a promising avenue for future research concerns the principled adjustment of the maximum entropy temperature coefficient ($\alpha$). In planning-driven MBRL, the sampling is dictated by the planner's search, yielding an intricate, non-trivial relationship between the planner's intrinsic exploration entropy and the network policy's entropy. Developing a automatic tuning mechanism for $\alpha$ based on these entropy signals is key to stabilizing the inherent tension between exploration and exploitation—a challenge that continues to define the frontier of effective RL algorithm design.
\section{Acknowledgment}
This study is supported by the Tsinghua University-Toyota Joint Research Center for AI Technology of Automated Vehicle, Beijing Natural Science Foundation (L257002), and SunRisingAI Lab.

\bibliographystyle{plain}
\bibliography{ref}

\newpage
\appendix

\section{Theoretical Analysis}
\subsection{Useful Lemmas}

\begin{lemma}[Triangle Inequality]
\label{lemma_1}
For any vectors $x, y \in \mathbb{R}^n$, the triangle inequality states that the norm of the sum of two vectors is less than or equal to the sum of their norms:
\[
\|x + y\| \leq \|x\| + \|y\|,
\]
where $\|\cdot\|$ denotes the standard Euclidean norm.
\end{lemma}

\begin{lemma}[Pinsker’s Inequality]
\label{lemma_2}

Let $p$ and $q$ be two probability distributions over a measurable space $(\mathcal{X}, \mathcal{F})$. Denote the total variation distance between $p$ and $q$ as
\[
    \mathrm{TV}(p, q) := \sup_{A \in \mathcal{F}} |p(A) - q(A)| = \frac{1}{2} \int_{\mathcal{X}} |p(x) - q(x)| \, dx,
\]
where $p$ and $q$ are the probability density functions of $p$ and $q$, respectively. The forward Kullback–Leibler (KL) divergence from $p$ to $q$ is defined as
\[
    \mathrm{KL}(p \| q) := \int_{\mathcal{X}} p(x) \log \frac{p(x)}{q(x)} \, dx.
\]
Then the total variation distance is upper bounded by the square root of the forward KL divergence:
\[
    \mathrm{TV}(p, q) \le \sqrt{\tfrac{1}{2}\,\mathrm{KL}(p \| q)}.
\]
\end{lemma}

\begin{lemma}[Total Variation Bound on Expectation Difference]
\label{lemma_3}
Let \( p \) and \( q \) be two probability densities over a common measurable space \( \mathcal{X} \), and let \( f: \mathcal{X} \to \mathbb{R} \) be a measurable function such that \( \|f\|_\infty = \sup_{x \in \mathcal{X}} |f(x)| < \infty \). Then the difference in expectations is bounded by
\[
\left| \mathbb{E}_{x \sim p}[f(x)] - \mathbb{E}_{x \sim q}[f(x)] \right| \le 2 \|f\|_\infty \cdot \mathrm{TV}(p, q),
\]
\end{lemma}

\begin{proof}
We begin by expressing the difference in expectations:
\[
\left| \mathbb{E}_{p}[f(x)] - \mathbb{E}_{q}[f(x)] \right| = \left| \int_{\mathcal{X}} f(x)(p(x) - q(x)) \, dx \right|.
\]
By the triangle inequality,
\[
\left| \int_{\mathcal{X}} f(x)(p(x) - q(x)) \, dx \right| \le \int_{\mathcal{X}} |f(x)||p(x) - q(x)| \, dx.
\]
Using the fact that \( |f(x)| \le \|f\|_\infty \), we obtain
\[
\int_{\mathcal{X}} |f(x)||p(x) - q(x)| \, dx \le \|f\|_\infty \int_{\mathcal{X}} |p(x) - q(x)| \, dx.
\]
Recalling the definition of total variation distance,
\[
\int_{\mathcal{X}} |p(x) - q(x)| \, dx = 2 \, \mathrm{TV}(p, q),
\]
we conclude that
\[
\left| \mathbb{E}_{p}[f(x)] - \mathbb{E}_{q}[f(x)] \right| \le 2 \|f\|_\infty \cdot \mathrm{TV}(p, q).
\]
\end{proof}

\begin{lemma}[Concentration of Gaussian Policy Samples]
\label{lemma_4}
Let \( a_\pi \sim \mathcal{N}(\mu_\pi, \Sigma_\pi) \) be a sample from a multivariate Gaussian distribution with mean \( \mu_\pi \in \mathbb{R}^d \) and covariance matrix \( \Sigma_\pi \in \mathbb{R}^{d \times d} \). Define
\[
\Lambda(\Sigma_\pi) := \sup_{\|v\|_2 = 1} v^\top \Sigma_\pi v
\]
as the largest directional variance (i.e., the spectral norm of \( \Sigma_\pi \)). Then, for any \( \delta \in (0, 1) \), with probability at least \( 1 - \delta \),
\[
\| a_\pi - \mu_\pi \|_2 \le \sqrt{2 \Lambda(\Sigma_\pi) \log \tfrac{1}{\delta}}.
\]
\end{lemma}

\begin{proof}
Let \( z \sim \mathcal{N}(0, I_d) \), and let \( L \) be a matrix such that \( \Sigma_\pi = L L^\top \). Then the policy sample can be written as \( a_\pi = \mu_\pi + L z \), and we have
\[
\| a_\pi - \mu_\pi \|_2^2 = \| L z \|_2^2 = z^\top \Sigma_\pi z.
\]
Using a standard concentration bound for sub-Gaussian quadratic forms (e.g., the Laurent–Massart inequality), for any \( \delta \in (0,1) \),
\[
\Pr\left( z^\top \Sigma_\pi z \ge \mathbb{E}[z^\top \Sigma_\pi z] + 2 \Lambda(\Sigma_\pi) \log \tfrac{1}{\delta} \right) \le \delta.
\]
Since \( \mathbb{E}[z^\top \Sigma_\pi z] = \mathrm{Tr}(\Sigma_\pi) \), we obtain
\[
\Pr\left( \| a_\pi - \mu_\pi \|_2^2 \ge \mathrm{Tr}(\Sigma_\pi) + 2 \Lambda(\Sigma_\pi) \log \tfrac{1}{\delta} \right) \le \delta.
\]
By omitting the trace term, we get a looser but simpler bound:
\[
\| a_\pi - \mu_\pi \|_2 \le \sqrt{2 \Lambda(\Sigma_\pi) \log \tfrac{1}{\delta}}, \quad \text{with probability at least } 1 - \delta.
\]
\end{proof}

\begin{lemma}[Mean Bound via KL Between Gaussians]
\label{lemma_5}
Let \( p = \mathcal{N}(\mu_p, \Sigma_p) \) and \( q = \mathcal{N}(\mu_q, \Sigma_q) \) be two multivariate Gaussian distributions of dimension \( d \). Then,
\[
    \mathrm{KL}(p \| q) \ge \frac{1}{2} \| \mu_p - \mu_q \|_{\Sigma_q^{-1}}^2.
\]
Consequently, if \( \mathrm{KL}(p \| q) \le \varepsilon \), then
\[
    \| \mu_p - \mu_q \|_2 \le \sqrt{2 \varepsilon \Lambda(\Sigma_q)},
\]
where \( \Lambda(\Sigma_q) \) denotes the largest eigenvalue of \( \Sigma_q \).
\end{lemma}

\begin{proof}
The KL divergence between Gaussians is given by:
\[
\mathrm{KL}(p \| q) = \frac{1}{2} \left( \operatorname{tr}(\Sigma_q^{-1} \Sigma_p) + (\mu_q - \mu_p)^\top \Sigma_q^{-1} (\mu_q - \mu_p) - d + \log \frac{\det \Sigma_q}{\det \Sigma_p} \right).
\]
Dropping the non-negative trace and log-determinant terms, we obtain the lower bound:
\[
\mathrm{KL}(p \| q) \ge \frac{1}{2} \| \mu_p - \mu_q \|_{\Sigma_q^{-1}}^2.
\]
Applying the spectral norm inequality \( \| v \|_2^2 \le \Lambda(\Sigma_q) \| v \|_{\Sigma_q^{-1}}^2 \), we get:
\[
\| \mu_p - \mu_q \|_2^2 \le 2 \varepsilon \Lambda(\Sigma_q).
\]

This completes the proof.
\end{proof}



\subsection{Proof of Theorem 1}
\label{proof_theorem_1}
\begin{proof}
We aim to bound the difference between the expected returns under two policies \( \pi \) and \( \beta \), based on the divergence of their induced trajectory distributions.

By definition, the expected return under a policy \( \pi \) is:
\[
J(\pi) = \mathbb{E}_{\tau \sim \pi} \left[ \sum_{t=0}^{\infty} \gamma^t r(s_t, a_t) \right],
\]
where \( \tau = (s_0, a_0, s_1, a_1, \ldots) \) denotes a full trajectory, and similarly for $\beta$, we obtain
\[
J(\beta) = \mathbb{E}_{\tau \sim \beta} \left[ \sum_{t=0}^{\infty} \gamma^t r(s_t, a_t) \right],
\]

We compute the difference:
\[
|J(\beta) - J(\pi)| = \left| \sum_{t=0}^\infty \gamma^t \left( \mathbb{E}_{(s_t, a_t) \sim \beta} [r(s_t, a_t)] - \mathbb{E}_{(s_t, a_t) \sim \pi} [r(s_t, a_t)] \right) \right|.
\]

Using the triangle inequality in Lemma \ref{lemma_1}, this is upper bounded by
\[
\sum_{t=0}^\infty \gamma^t \left| \mathbb{E}_{(s_t, a_t) \sim \beta} [r(s_t, a_t)] - \mathbb{E}_{(s_t, a_t) \sim \pi} [r(s_t, a_t)] \right|.
\]

Assuming the reward function is uniformly bounded as \( |r(s,a)| \le R_{\max} \), Lemma~\ref{lemma_3} implies the following bound:
\[
\left| \mathbb{E}_{(s_t, a_t) \sim \beta} [r(s_t, a_t)] - \mathbb{E}_{(s_t, a_t) \sim \pi} [r(s_t, a_t)] \right| \le 2 R_{\max} \cdot \mathrm{TV}(p^\beta_t, p^\pi_t),
\]
where \( p^\beta_t \) and \( p^\pi_t \) denote the marginal distributions of \( (s_t, a_t) \) under policies \( \pi \) and \( \beta \), respectively.

Substituting this into the previous expression, we obtain:
\[
|J(\beta) - J(\pi)| \le 2 R_{\max} \sum_{t=0}^{\infty} \gamma^t \cdot \mathrm{TV}(p^\beta_t, p^\pi_t).
\]

Now, applying Pinsker’s inequality in  Lemma \ref{lemma_2}, we further have
\[
\mathrm{TV}(p^\beta_t, p^\pi_t) \le \sqrt{\frac{1}{2} \mathrm{KL}(p^\beta_t \| p^\pi_t)}.
\]

Assuming that at every timestep, the KL divergence is bounded as \( \mathrm{KL}(p^\beta_t \| p^\pi_t) \le \varepsilon \), we get:
\[
|J(\beta) - J(\pi)| \le 2 R_{\max} \sum_{t=0}^{\infty} \gamma^t \sqrt{\frac{\varepsilon}{2}} = \frac{2 R_{\max}}{1 - \gamma} \sqrt{\frac{\varepsilon}{2}} = \frac{R_{\max}}{1 - \gamma} \sqrt{2\varepsilon}.
\]

This completes the proof.
\end{proof}

\subsection{Proof of Theorem 2}
\label{proof_theorem_2}
\begin{proof}
Let $a_\pi \sim \pi(s)$ and $a_\beta \sim \beta(s)$, we begin by applying the Lipschitz continuity of \( Q(s,a) \):
\[
\left|   Q(s, a_\beta) - Q(s, a_\pi) \right| 
\le L_Q \cdot \| a_\beta - a_\pi \|_2 .
\]

The difference in the RHS is typically bounded by the forward KL divergence between these two distributions, i.e., $\| a_\beta - a_\pi \|_2 \le D(\epsilon)$, where $\epsilon$ is the forward KL divergence bound, i.e., $\mathrm{KL}(\beta \| \pi) \leq \epsilon$. 
This bound provides an upper limit on the discrepancy between the sampled actions from both policies, reflecting how much the policies differ in terms of their action distributions. 

We consider the case that
$
\beta(s) = \sum_{i=1}^K w_i \mathcal{N}(\mu_i, \Sigma_i),
$
where the mixture weights satisfy \( w_i \ge 0 \) and \( \sum_{i=1}^K w_i = 1 \).  
Policy \( \pi(s) = \mathcal{N}(\mu_\pi, \Sigma_\pi) \) is a Gaussian.  
Let \( \Lambda_i := \Lambda(\Sigma_i) \) denote the largest eigenvalue of each component covariance \( \Sigma_i \), and let \( \Lambda(\Sigma_\pi) \) denote the largest eigenvalue of \( \Sigma_\pi \).

Consider the decomposition using triangle inequality in Lemma~\ref{lemma_1}:
\[
\| a_\beta - a_\pi \|_2 \le \max_i \big( \| a_\beta - \mu_i \|_2 + \| \mu_i - \mu_\pi \|_2 \big) + \| a_\pi - \mu_\pi \|_2.
\]

\textbf{(1) Sampling deviation of GMM:}  
Conditional on sampling component \( i \sim w \), the action sample \( a_\beta \sim \mathcal{N}(\mu_i, \Sigma_i) \). For each component, we apply Lemma~\ref{lemma_4} to obtain the following bound:
\[
\| a_\beta - \mu_i \|_2 \le \sqrt{2 \Lambda_i \log \tfrac{K}{\delta}} \quad \text{with probability at least } 1 - \delta.
\]

\textbf{(2) Mean shift of GMM components:}  
We now bound the deviation between the GMM component means \( \mu_i \) and the target policy mean \( \mu_\pi \). From the definition of KL divergence between Gaussians:
\[
\mathrm{KL}\left( \mathcal{N}(\mu_i, \Sigma_i) \| \mathcal{N}(\mu_\pi, \Sigma_\pi) \right) \ge \frac{1}{2} \| \mu_i - \mu_\pi \|_{\Sigma_\pi^{-1}}^2.
\]
Therefore, if we know that the total mixture KL satisfies \( \mathrm{KL}(\beta \| \pi) \le \varepsilon \), then by invoking Lemma \ref{lemma_5}, we have: for each component \( i \), it must be that
\[
\| \mu_i - \mu_\pi \|_{\Sigma_\pi^{-1}}^2 \le 2 \cdot \frac{\varepsilon}{w_i} \quad \Rightarrow \quad 
\| \mu_i - \mu_\pi \|_2 \le \sqrt{2 \varepsilon \Lambda(\Sigma_\pi) / w_i}.
\]
Hence, plugging this into the deviation bound, we now obtain:
\[
\| a_\beta - a_\pi \|_2 \le \max_i \Big( \sqrt{2 \Lambda_i \log \tfrac{K}{\delta}} + \sqrt{2 \varepsilon \Lambda(\Sigma_\pi) / w_i} \Big) + \| a_\pi - \mu_\pi \|_2,
\]
with high probability at least \( 1 - \delta \).

\textbf{(3) Sampling deviation of \( \pi \):}  
As \( a_\pi \sim \mathcal{N}(\mu_\pi, \Sigma_\pi) \), by invoking Lemma~\ref{lemma_4} again, we have:
\[
\| a_\pi - \mu_\pi \|_2 \le \sqrt{2 \Lambda(\Sigma_\pi) \log \tfrac{1}{\delta}}.
\]

Putting everything together, with probability at least \( 1 - \delta \), we have:
\[
\| a_\beta - a_\pi \|_2 
\le \max_i \Big( \sqrt{2 \Lambda_i \log \tfrac{K}{\delta}} + \sqrt{2 \varepsilon \Lambda(\Sigma_\pi) / w_i} \Big) + 
    \sqrt{2 \Lambda(\Sigma_\pi) \log \tfrac{1}{\delta}}.
\]

Since both \( a_\beta \) and \( a_\pi \) are supported on the normalized action space \( [-1,1]^d \), the maximum possible distance is bounded by \( 2 \sqrt{d} \). Hence:
\[
\| a_\beta - a_\pi \|_2 \le 
\min \left( 2 \sqrt{d}, \
\max_i \Big( \sqrt{2 \Lambda_i \log \tfrac{K}{\delta}} + \sqrt{2 \varepsilon \Lambda(\Sigma_\pi) / w_i} \Big) + 
    \sqrt{2 \Lambda(\Sigma_\pi) \log \tfrac{1}{\delta}} 
\right).
\]

\textbf{Conclusion.}  
For any \( \delta \in (0,1) \), with probability at least \( 1 - \delta \), we obtain the bound:
\[
D(\varepsilon) \le \min \left( 2 \sqrt{d}, \
\max_i \Big( \sqrt{2 \Lambda_i \log \tfrac{K}{\delta}} + \sqrt{2 \varepsilon \Lambda(\Sigma_\pi) / w_i} \Big) + 
    \sqrt{2 \Lambda(\Sigma_\pi) \log \tfrac{1}{\delta}} 
\right).
\]
This completes the proof.
\end{proof}

\newpage
\section{Environmental Details
}
\subsection{Benchmark Introduction}
\label{benchmark}
\textbf{DeepMind Control Suite.} We evaluate on the 7 most challenging tasks involving the \texttt{dog} and \texttt{humanoid} agents. These tasks fall into two categories:  
(1) \emph{Standing tasks}, where the agent must maintain upright balance, and  
(2) \emph{Moving tasks}, which additionally require the agent to move at a target velocity.
For moving tasks, the reward is defined as the product of the standing reward and the forward velocity reward, i.e., 
$
\textbf{\textit{Reward}} = (\textbf{\textit{Standing reward}}) \times (\textbf{\textit{Forward velocity reward}}).
$


\begin{figure}[h!]
    \begin{minipage}[b]{0.235\textwidth}
        \centering
        \includegraphics[width=\linewidth]{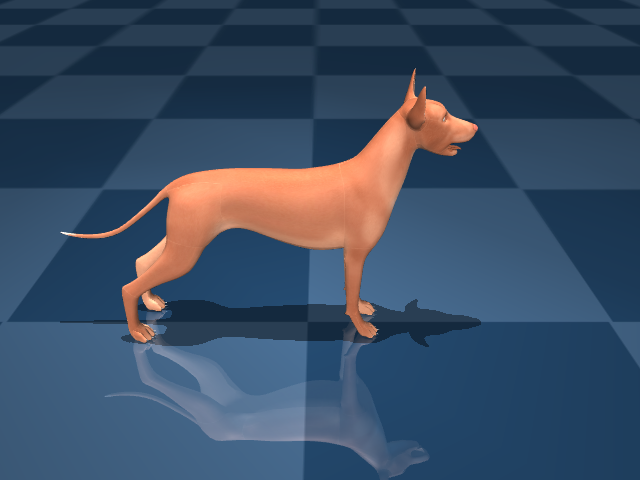}
        \caption{Dog}
        \label{fig_dog_run}
    \end{minipage}%
    \quad
    \begin{minipage}[b]{0.235\textwidth}
        \centering
        \includegraphics[width=\linewidth]{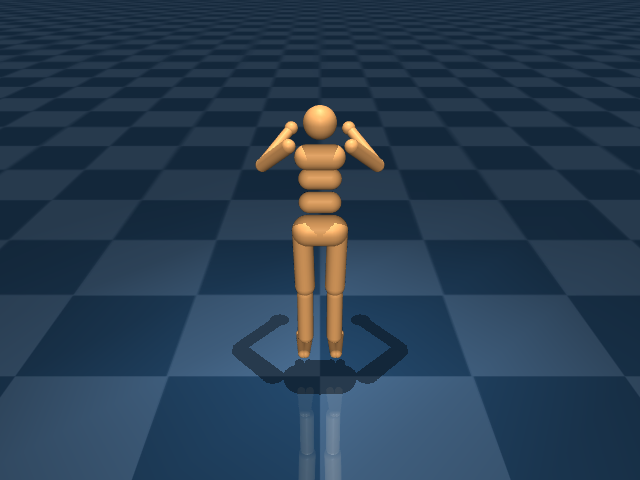}
        \caption{Humanoid}
        \label{fig_humanoid_run}
    \end{minipage}%
    \quad
    \begin{minipage}[b]{0.48\textwidth}
        \textbf{\textit{Standing reward}}: Encourages the agent to maintain an upright posture.\\
        
        \textbf{\textit{Forward velocity reward}}: Ensures the agent moves at the target speed (\SI{1}{\meter/\second} for dog-walk, \SI{3}{\meter/\second} for dog-trot, \SI{9}{\meter/\second} for dog-run, \SI{1}{\meter/\second} for humanoid-walk and \SI{10}{\meter/\second} for humanoid-run).\\
        \vspace{5pt}
    \end{minipage}
\end{figure}

\textbf{Humanoid Bench}. We consider 7 typical locomotion tasks involving a Unitree H1hand robot. This robot is initialized to a standing position, with random noise added to all joint positions during each episode reset. Their specific goals are presented below.

\begin{figure}[h!]
    \begin{minipage}[b]{0.25\textwidth}
        \centering
        \includegraphics[width=\linewidth]{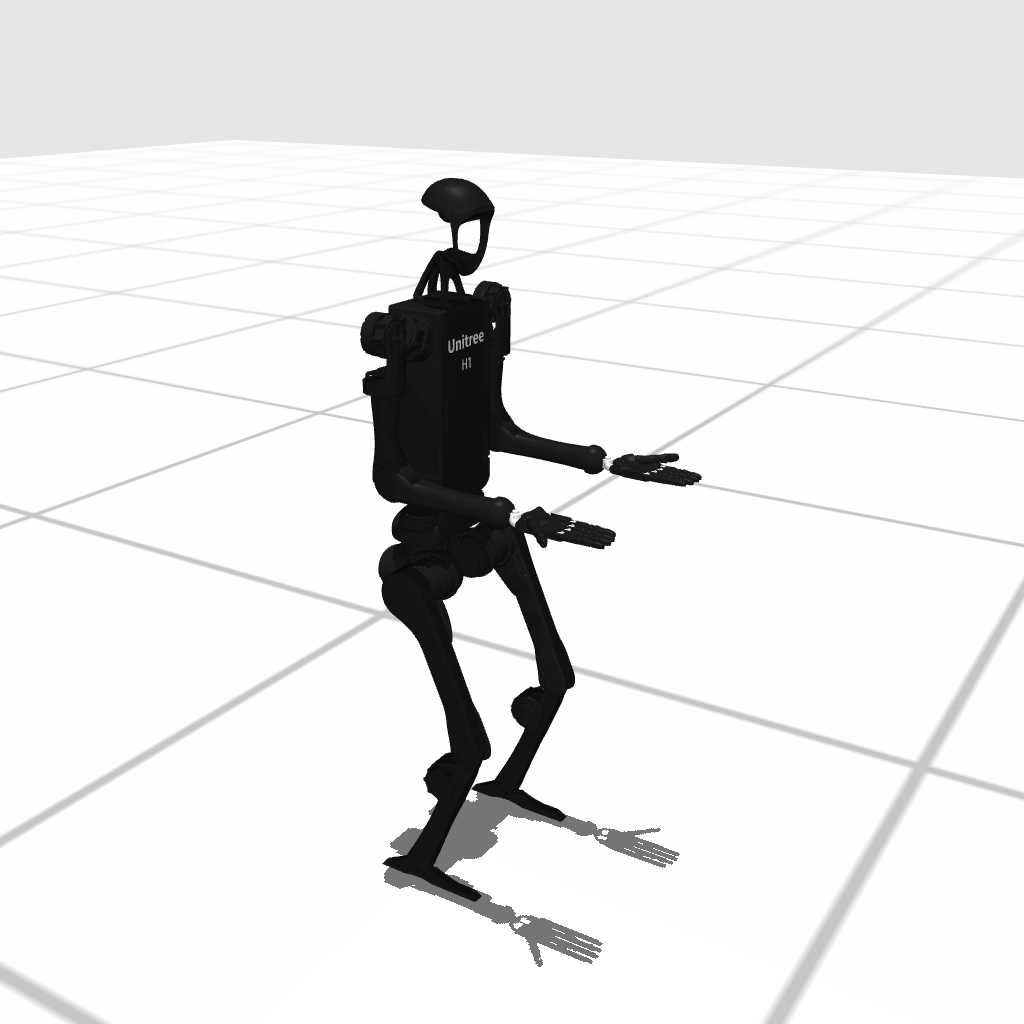}
        \caption{Stand}
        \label{fig_stand}
    \end{minipage}%
    \quad
    \begin{minipage}[b]{0.5\textwidth}
        \textbf{\textit{Objective.}} Maintain a standing pose.\\
        
        
        \textbf{\textit{Reward}}: $
R(s, a) = \texttt{stable} \times (0.5 \times \texttt{still}_x + 0.5 \times \texttt{still}_y),
$
where the \texttt{still} terms penalize non-zero velocities to encourage stationary balance. \texttt{stable} favors maintaining a stable and energy-efficient standing status.\\

        \textbf{\textit{Termination.}}  1000 steps, or when $z_\text{pelvis} < 0.2$. \\
    \end{minipage}
\end{figure}
\begin{figure}[h!]
    \begin{minipage}[b]{0.25\textwidth}
        \centering
        \includegraphics[width=\linewidth]{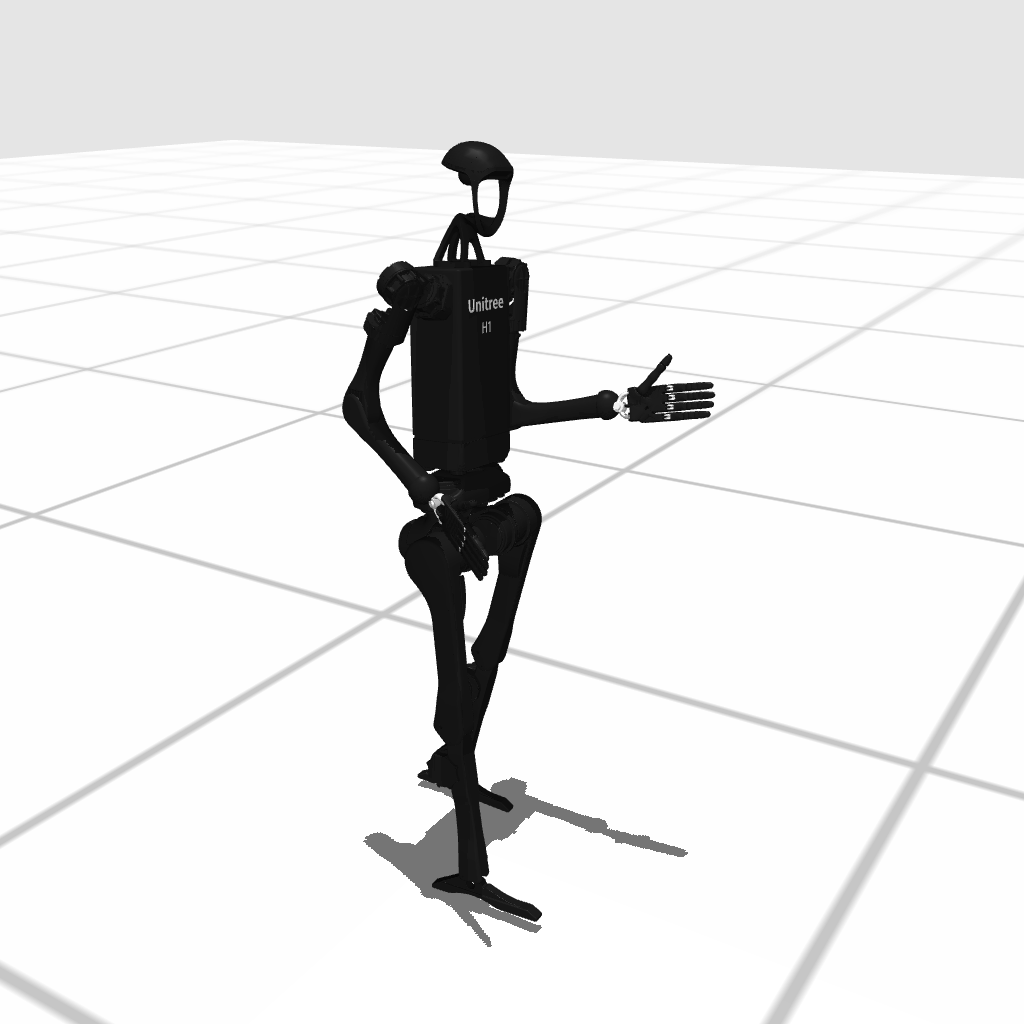}
        \caption{Walk}
        \label{fig_walk}
    \end{minipage}%
    \quad 
    \begin{minipage}[b]{0.5\textwidth}
        
        \textbf{\textit{Objective.}} Keep forward velocity close to \SI{1}{\meter/\second} without falling to the ground.\\
        
        
        \textbf{\textit{Reward}}: $R(s, a) = \texttt{stable} \times \texttt{tol}(v_x, (1, \infty), 1)$, where \texttt{tol} encourages the agent to maintain a forward velocity $v_x$ above $1\,\mathrm{m/s}$, thereby promoting  low-speed locomotion. \\

        \textbf{\textit{Termination.}}  1000 steps, or when $z_\text{pelvis} < 0.2$.\\
    \end{minipage}
\end{figure}

\begin{figure}[h!]
    \begin{minipage}[b]{0.25\textwidth}
        \centering
        \includegraphics[width=\linewidth]{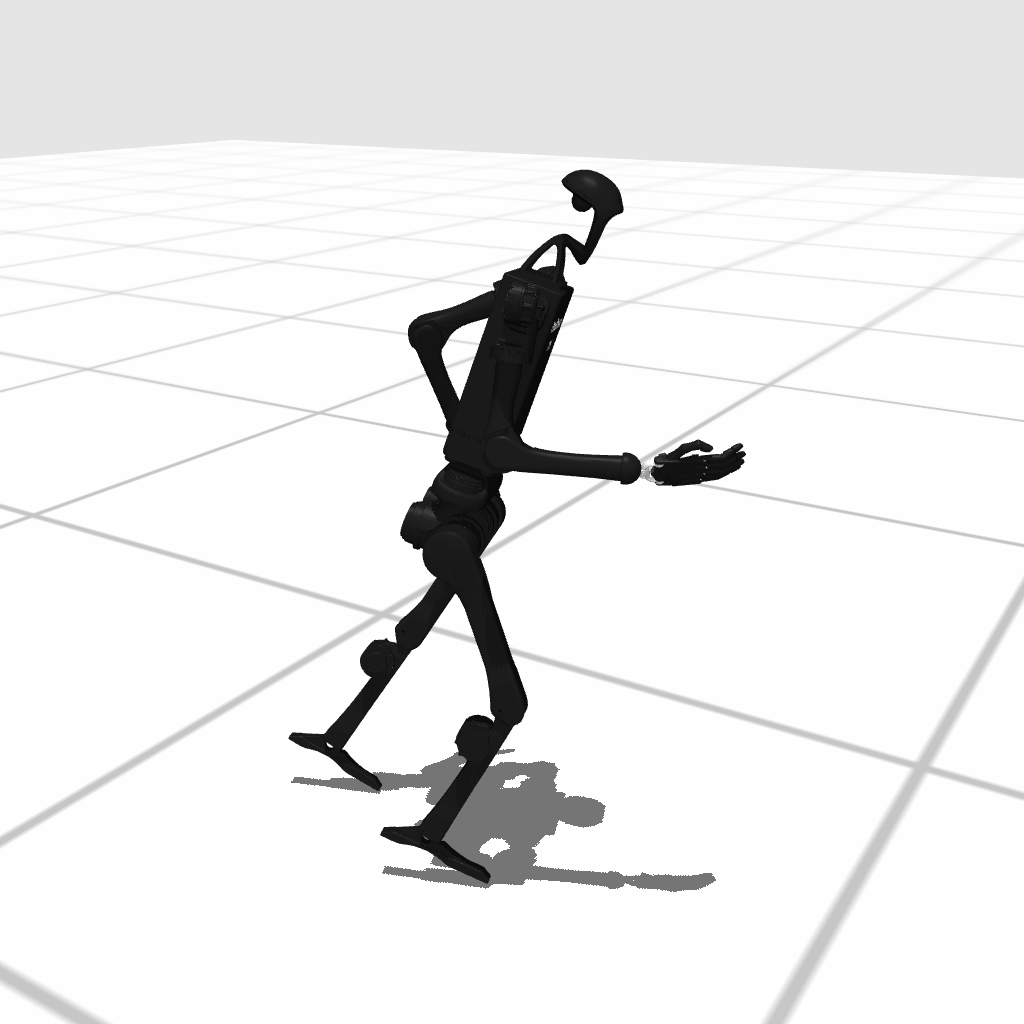}
        \caption{Run}
        \label{fig_run}
    \end{minipage}%
    \quad
    \begin{minipage}[b]{0.5\textwidth}
        \textbf{\textit{Objective.}} Keep forward velocity close to \SI{5}{\meter/\second} without falling to the ground.\\
        
        
        \textbf{\textit{Reward}}: $R(s, a) = \texttt{stable} \times \texttt{tol}(v_x, (5, \infty), 5)$, where \texttt{tol} encourages the agent to maintain a forward velocity $v_x$ above $5\,\mathrm{m/s}$, thereby promoting  high-speed locomotion. \\

        \textbf{\textit{Termination.}} 1000 steps, or when $z_\text{pelvis} < 0.2$.\\
    \end{minipage}
\end{figure}


\newpage
\begin{figure}[h!]
    \begin{minipage}[b]{0.25\textwidth}
        \centering
        \includegraphics[width=\linewidth]{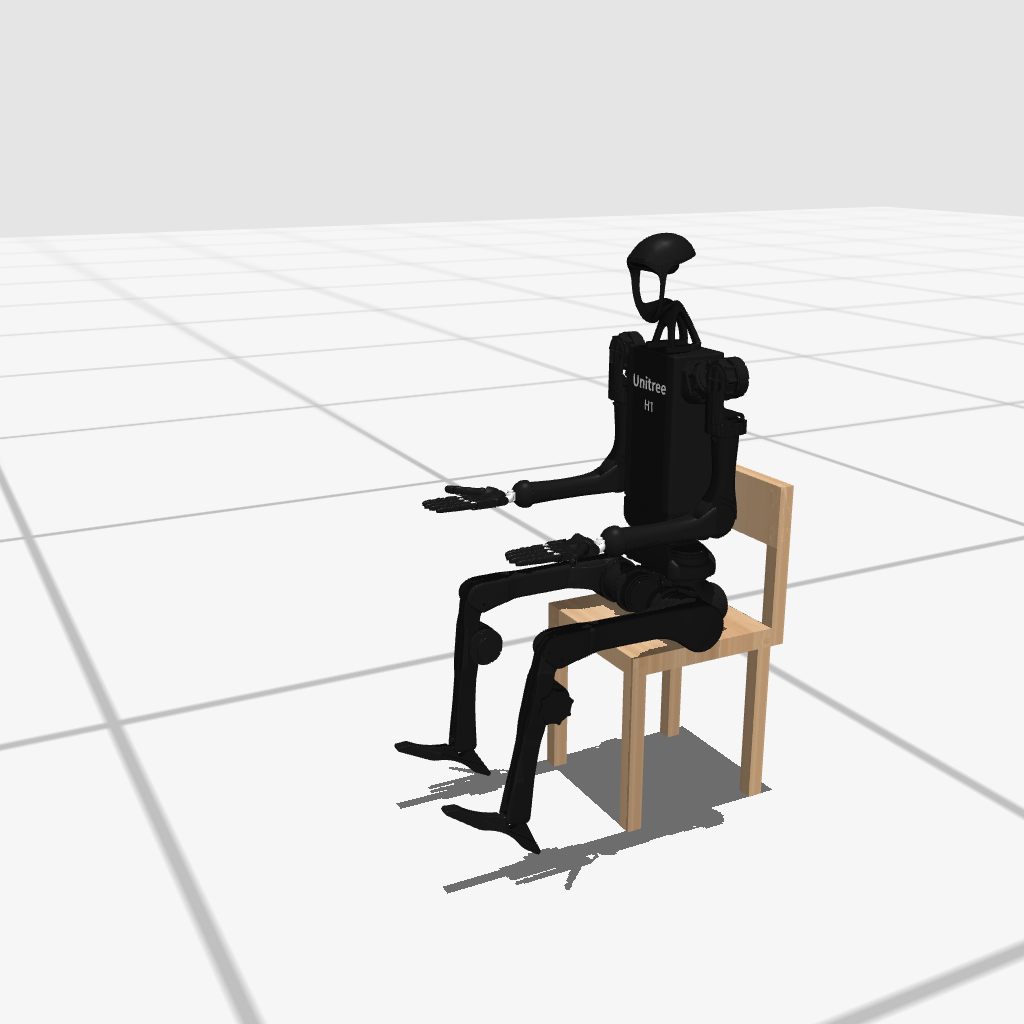}
        \caption{Sit}
        \label{fig_sit}
    \end{minipage}%
    \quad
    \begin{minipage}[b]{0.5\textwidth}
        \textbf{\textit{Objective.}} Sit onto a chair situated closely behind.\\
        
        \textbf{\textit{Reward}}: $R(s, a) = (0.5 \cdot \texttt{sitting\_z} + 0.5 \cdot \texttt{sitting\_x} \cdot \texttt{sitting\_y}) \times \texttt{upright} \times \texttt{posture} \times e  \times \texttt{mean}(\texttt{still\_x}, \texttt{still\_y})$,  where $e$ is an energy penalty term, $\texttt{sitting\_x}$, $\texttt{sitting\_y}$, and $\texttt{sitting\_z}$ measure the robot's positional tolerance relative to the chair. \\

        \textbf{\textit{Termination.}} 1000 steps, or when $z_\text{pelvis} < 0.5$.\\
    \end{minipage}
\end{figure}


\begin{figure}[h!]
    \begin{minipage}[b]{0.25\textwidth}
        \centering
        \includegraphics[width=\linewidth]{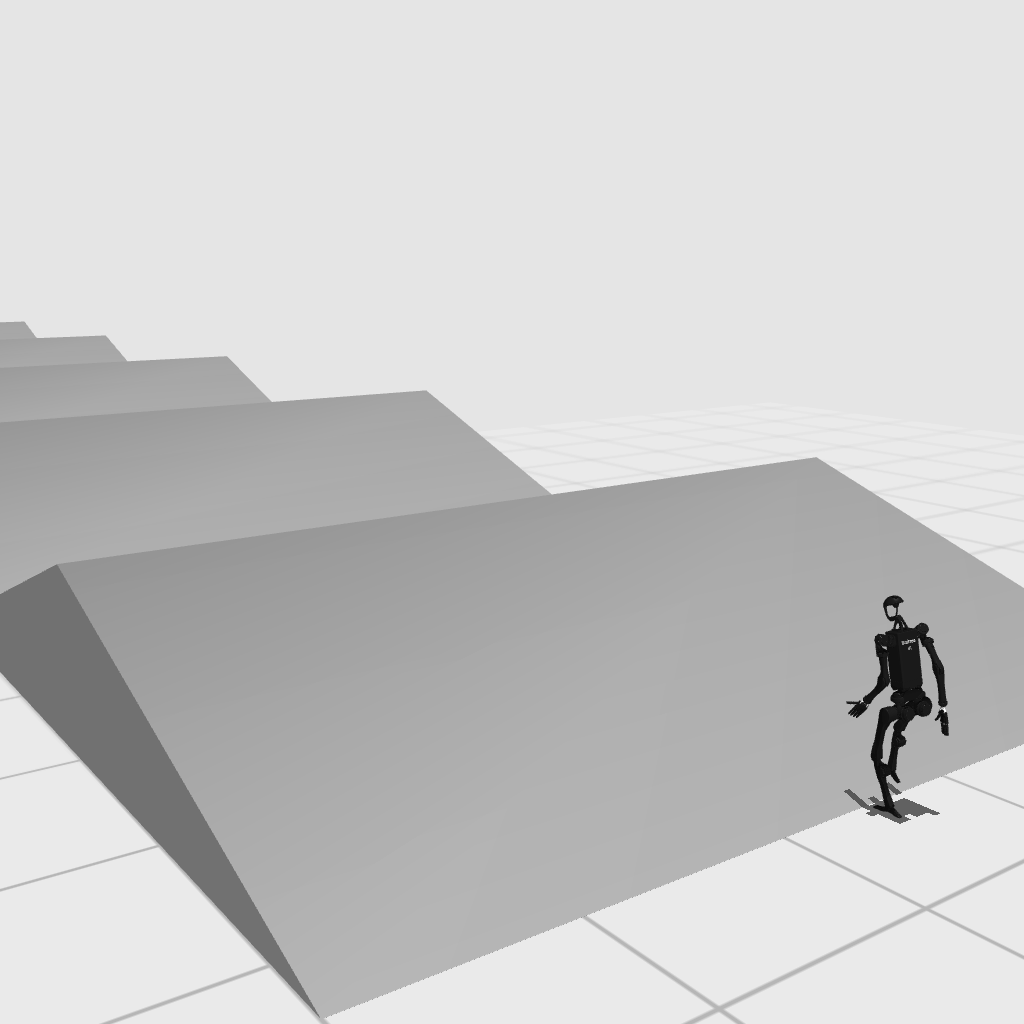}
        \caption{Slide}
        \label{fig_slide}
    \end{minipage}%
    \quad
    \begin{minipage}[b]{0.5\textwidth}
        \textbf{\textit{Objective.}} Walk over an iterating sequence of upward and downward slides at \SI{1}{\meter/\second}.\\
        
        

        \textbf{\textit{Reward}}: $
R(s, a) = e \times \texttt{tol}(v_x, (1, +\infty), 1) \times \texttt{upright} \times \bigl(\texttt{foot\_left} \times \texttt{foot\_right}\bigr),
$
where $\texttt{foot\_left}$ and $\texttt{foot\_right}$ measure the vertical distance between the head and left/right foot respectively, ensuring proper foot positioning.\\

        \textbf{\textit{Termination.}} 1000 steps, or when $z_\text{proj} < 0.6$. \\
    \end{minipage}
\end{figure}


\begin{figure}[h!]
    \begin{minipage}[b]{0.25\textwidth}
        \centering
        \includegraphics[width=\linewidth]{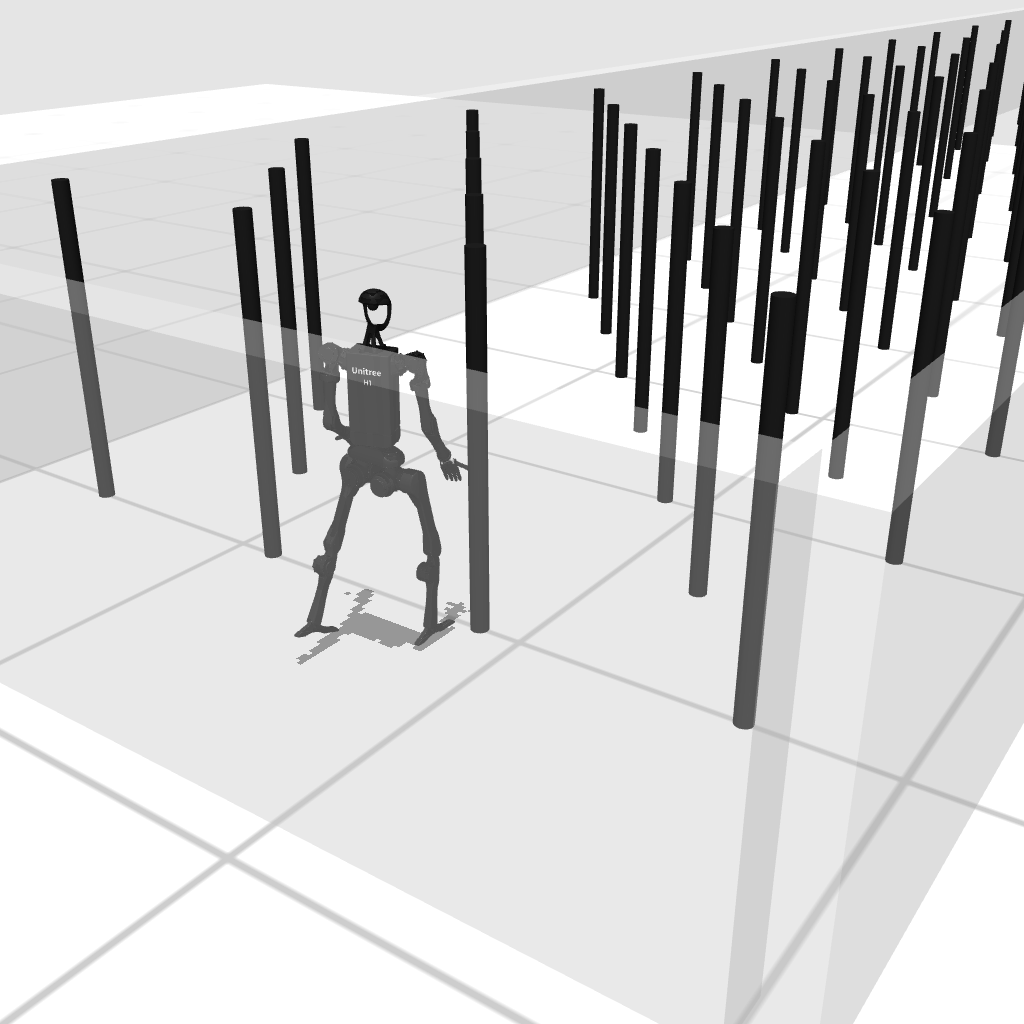}
        \caption{Pole}
        \label{fig_pole}
    \end{minipage}%
    \quad
    \begin{minipage}[b]{0.5\textwidth}
        \textbf{\textit{Objective.}} Travel forward over a dense forest of high thin poles, without colliding with them.\\
        
        
        \textbf{\textit{Reward}}: 
$
R(s, a) = \gamma_{\text{collision}} \times \bigl(0.5 \times \text{stable} + 0.5 \times \texttt{tol}(v_x, (1, +\infty), 1)\bigr),
$
where the collision penalty $\gamma_{\text{collision}}$ equals 0.1 if the robot collides with a pole, and 1 otherwise. \\
        
        \textbf{\textit{Termination.}} 1000 steps, or when $z_\text{pelvis} < 0.6$.\\
    \end{minipage}
\end{figure}


\begin{figure}[h!]
    \begin{minipage}[b]{0.25\textwidth}
        \centering
        \includegraphics[width=\linewidth]{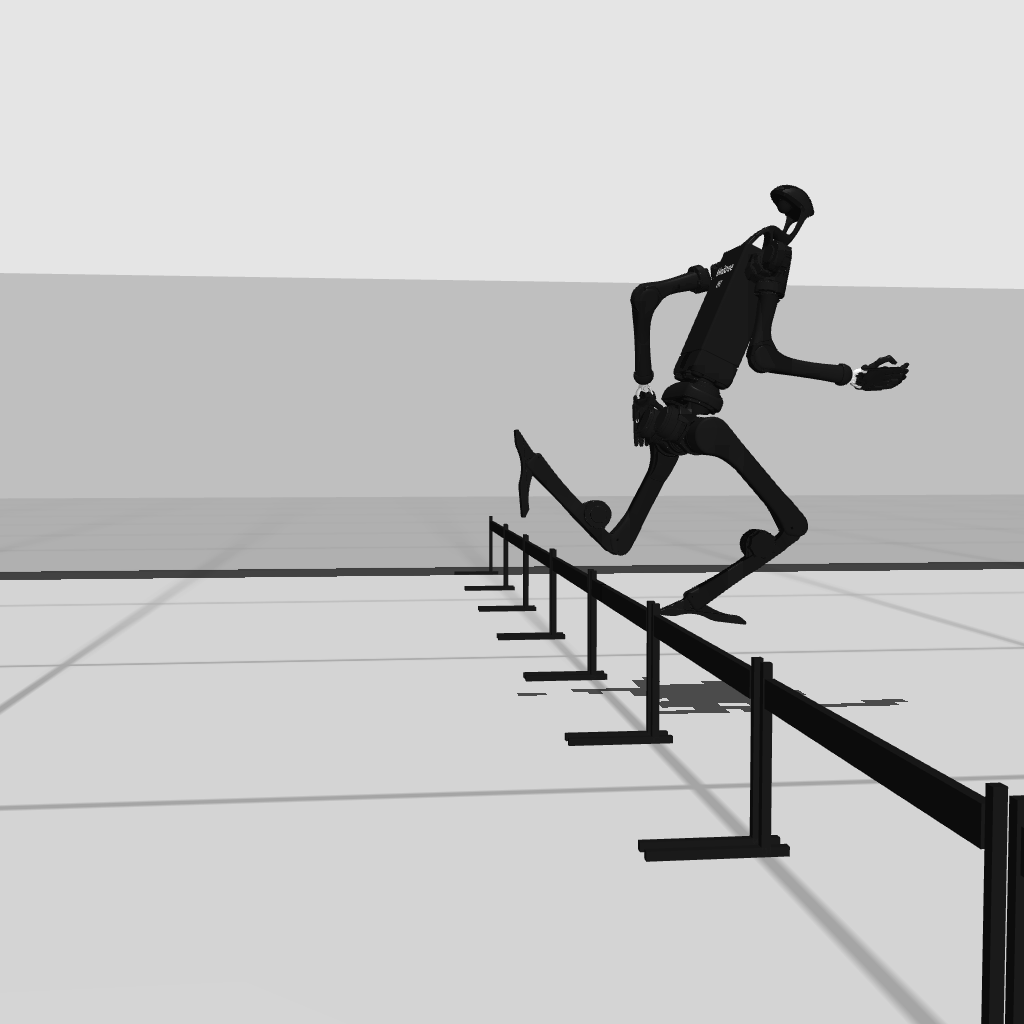}
        \caption{Hurdle}
        \label{fig_hurdle}
    \end{minipage}%
    \quad
    \begin{minipage}[b]{0.5\textwidth}
        \textbf{\textit{Objective.}} Keep forward velocity close to \SI{5}{\meter/\second} without falling to the ground.\\
        
        
        \textbf{\textit{Reward}}: 
$
R(s, a) = \text{stable} \times \texttt{tol}(v_x, (5, \infty), 5) \times \gamma_{\text{collision}},
$
which penalizes colliding with hurdle. \\

        \textbf{\textit{Termination.}}  1000 steps.\\
        \vspace{10pt}
    \end{minipage}
\end{figure}


\subsection{Reproducibility Statement \&  Detailed Hyperparameters}
\label{reproduce}
\label{hyperparameters}

We base all our experiments on the released official TD-MPC2 codebase {\url{https://github.com/nicklashansen/tdmpc}. We adopt their hyperparameter settings without additional tuning and use the same configuration across all previously demonstrated tasks. The details are listed in Table \ref{tab_hyperparams}. Our core algorithm file and video demos for the most challenging hurdle, pole and slide tasks are accessible at \url{https://anonymous.4open.science/r/NeurIPS_BOOM-C587}.

In this paper, we evaluate each algorithm for each
tasks over three random seeds. The CPU used is the AMD Ryzen
Threadripper 3960X 24-Core Processor, and the GPU used is NVIDIA GeForce RTX 3090Ti. Taking
\textit{Dog-run} task in the DMC Suite as an example, the time taken to train 2M iterations is around 50 hours.

For SAC and DreamerV3, we report the baseline results released by TD-MPC2, which are obtained from the official repositories of  {\url{https://github.com/denisyarats/pytorch_sac}} and  {\url{https://github.com/danijar/dreamerv3}}, respectively. For BMPC, we use the official implementation {\url{https://github.com/wertyuilife2/bmpc}} and align the settings such as number of iterations and evaluation protocol for a fair comparison.

\begin{table}[h]
\caption{Hyperparameter settings. }
\label{tab_hyperparams}
\vskip 0.15in
\begin{center}
\begin{small}
\begin{tabular}{l l | l l}
\toprule
\textbf{Hyperparameter}       & \textbf{Value}   & \textbf{Hyperparameter} & \textbf{Value} \\ 
\midrule
\multicolumn{4}{l}{\textbf{Training}} \\ \midrule
Learning rate                 & $3 \times 10^{-4}$  & Target network update rate    & 0.5       \\
Encoder learning rate                 & $1 \times 10^{-4}$  & Discount factor ($\gamma$)  & 0.99       \\
Sample batch size                    &  1 & Gradient Clipping Norm        & 20       \\
Replay batch size  & 256  & Optimizer                     & Adam  \\
Buffer size                   & 1\_000\_000  &  Loss norm & Moving $(5\%, 95\%)$ \\
Steps                      & 2\_000\_000  & Sampling                      & Uniform  \\
\midrule
\multicolumn{4}{l}{\textbf{World Model}} \\ \midrule
Reward loss coefficient ($c_r$)   & 0.1 & Dynamics loss coefficient ($c_f$)  & 20 \\
Value loss coefficient ($c_q$)  & 0.1  & Value functions esemble  & 5 \\
Number of value bins  & 101 & Warmup steps  & 5000 \\
\midrule
\multicolumn{4}{l}{\textbf{Planner}} \\ \midrule
MPPI Iterations     & 6 (8 if $\| \mathcal{A} \|$ > 20) & Minimum planner std  & 0.05 \\
Population size     & 512 & Maximum planner std  & 2  \\
Number of elites     & 64 & Horizon & 3 \\
Policy prior samples  & 24 & & \\
\midrule
\multicolumn{4}{l}{\textbf{Actor}} \\ \midrule
Minimum policy log std   & -10 & Entropy coefficient ($\alpha$) &  $1 \times 10^{-4}$  \\
Maximum policy log std   & 2  & \\
\midrule
\multicolumn{4}{l}{\textbf{Architecture (around 5M parameters in total)}} \\ \midrule
Encoder layers                & 2 & Latent space dimension        & 512 \\
Encoder dimension             & 256 & Task embedding dimension      & 96 \\
MLP hidden layer dimension    & 512 & Q function drop out rate & 0.01 \\
MLP activation                & Mish & MLP Normalization             & LayerNorm \\
\bottomrule
\end{tabular}
\end{small}
\end{center}
\vskip -0.1in
\end{table}

\section{Supplemental Results}

\subsection{Illustrative Discussion: Forward KL vs. Reverse KL}
\label{compare_kl}
To empirically illustrate the differences between forward and reverse KL objectives in aligning an actor policy with a multimodal planner distribution, we create a toy 1D problem for fitting  a mixture of two Gaussians, which simulates a planner policy that captures multiple high-value regions. We initialize a unimodal Gaussian policy \( q(x) = \mathcal{N}(x\,|\,\mu,\ \sigma^2) \), and optimize its parameters by minimizing either the \textbf{forward KL} divergence \( \mathrm{KL}(p \| q) \) or the \textbf{reverse KL} divergence \( \mathrm{KL}(q \| p) \). 

As shown in Figure~\ref{fig:kl}, the policy trained with \textbf{forward KL} consistently adjusts its mean and variance to cover both modes of the planner distribution, demonstrating a \emph{mode-covering} behavior. In contrast, the \textbf{reverse KL} objective tends to converge to only one mode of the distribution, often ignoring the global structure. This highlights its \emph{mode-seeking} nature and reveals the risk of it aligning with a \emph{suboptimal peak}, potentially converging to a low-value region.

\begin{figure}[h!]
    \centering
    \includegraphics[width=0.4\linewidth]{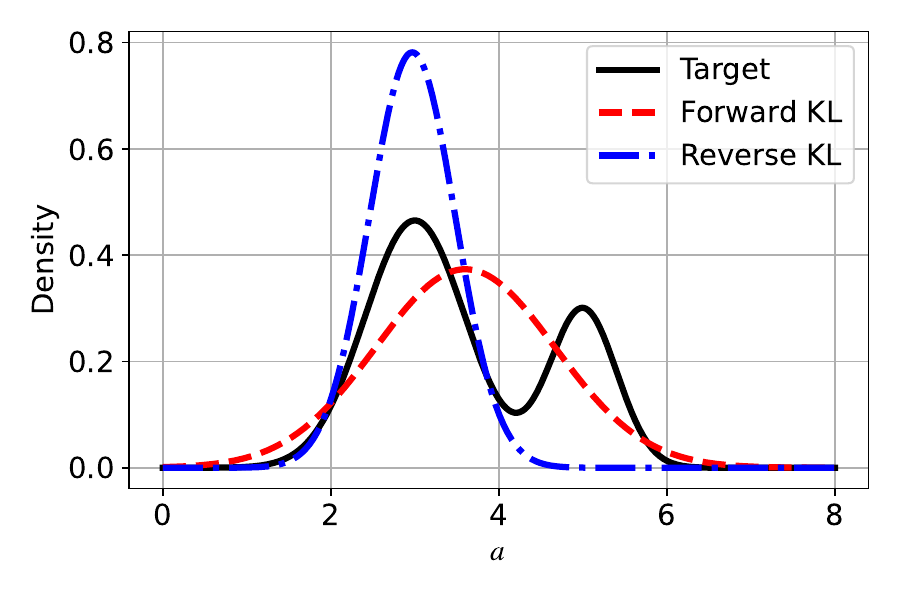}
    \caption{Comparison of forward KL and reverse KL on a fitting task. This example highlights a key issue: when the planner is multimodal but the policy is unimodal, \textbf{reverse KL may fail to sufficiently cover high-density regions of the target distribution}, still resulting in significant mismatch and poor performance. In contrast, \textbf{forward KL provides a more stable alignment strategy}, promoting broader coverage of the target distribution and preventing premature collapse.}
    \label{fig:kl}
\end{figure}

\subsection{Visualizations}
To demonstrate the effectiveness of BOOM in solving complex, high-dimensional locomotion tasks, we provide visualizations of policy control process on three of the most challenging benchmarks in the Humanoid Bench: \textbf{hurdle}, \textbf{pole}, and \textbf{slide} as shown in the following Figure \ref{fig:video_frames}. These tasks require precise coordination across many degrees of freedom, long-horizon reasoning, and dynamic interaction with many objects. The visualization showcase that BOOM not only achieves task success but also learns robust behaviors, highlighting its strong  capabilities in difficult control scenarios.

\begin{figure}[h]
  \centering
  \begin{subfigure}[t]{0.242\textwidth}
    \centering
    \includegraphics[width=\linewidth]{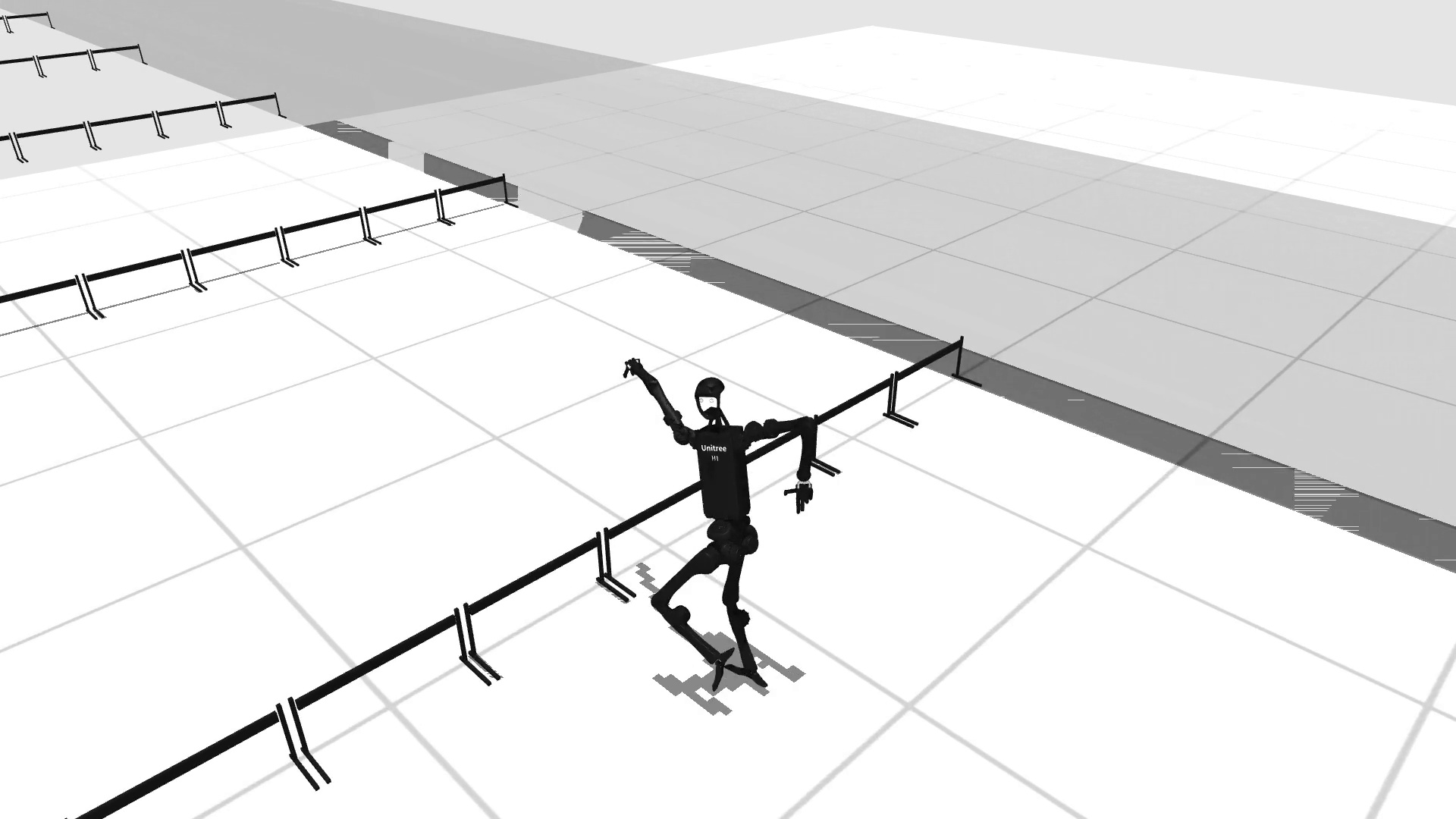}
    \caption{Frame 131}
  \end{subfigure}
  \begin{subfigure}[t]{0.242\textwidth}
    \centering
    \includegraphics[width=\linewidth]{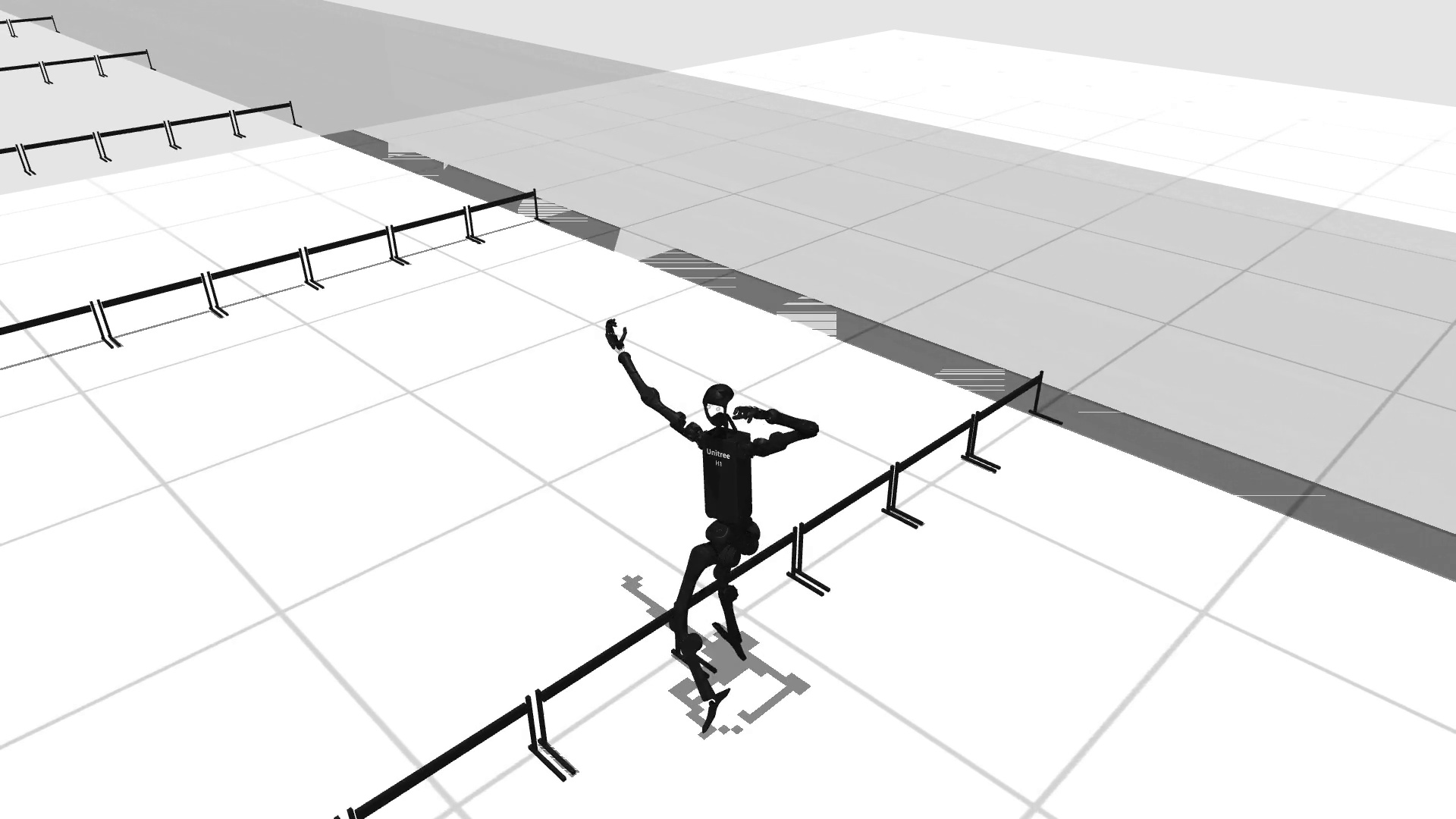}
    \caption{Frame 141}
  \end{subfigure}
  \begin{subfigure}[t]{0.242\textwidth}
    \centering
    \includegraphics[width=\linewidth]{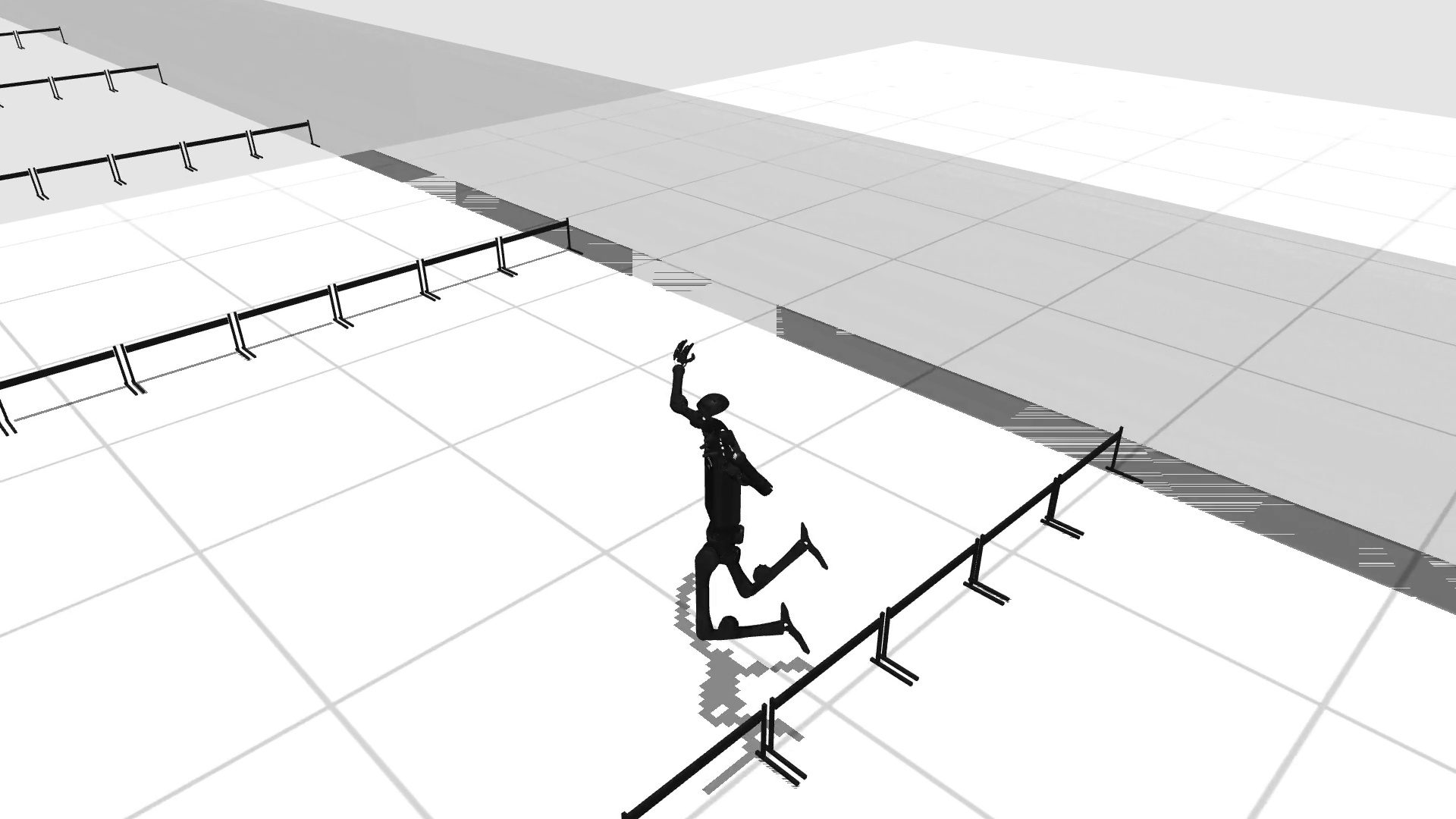}
    \caption{Frame 151}
  \end{subfigure}
  \begin{subfigure}[t]{0.242\textwidth}
    \centering
    \includegraphics[width=\linewidth]{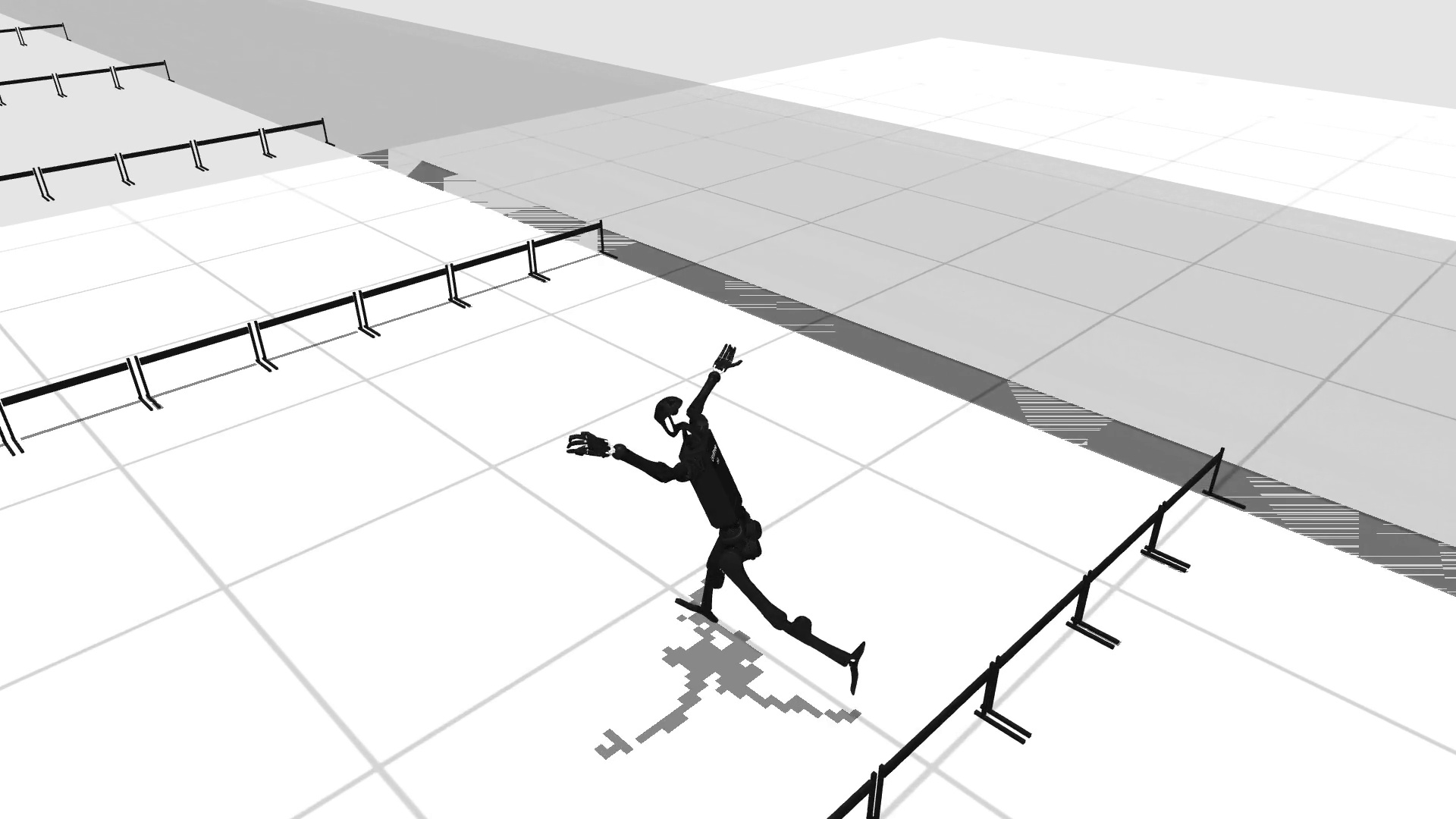}
    \caption{Frame 161}
  \end{subfigure}
    \begin{subfigure}[t]{0.242\textwidth}
    \centering
    \includegraphics[width=\linewidth]{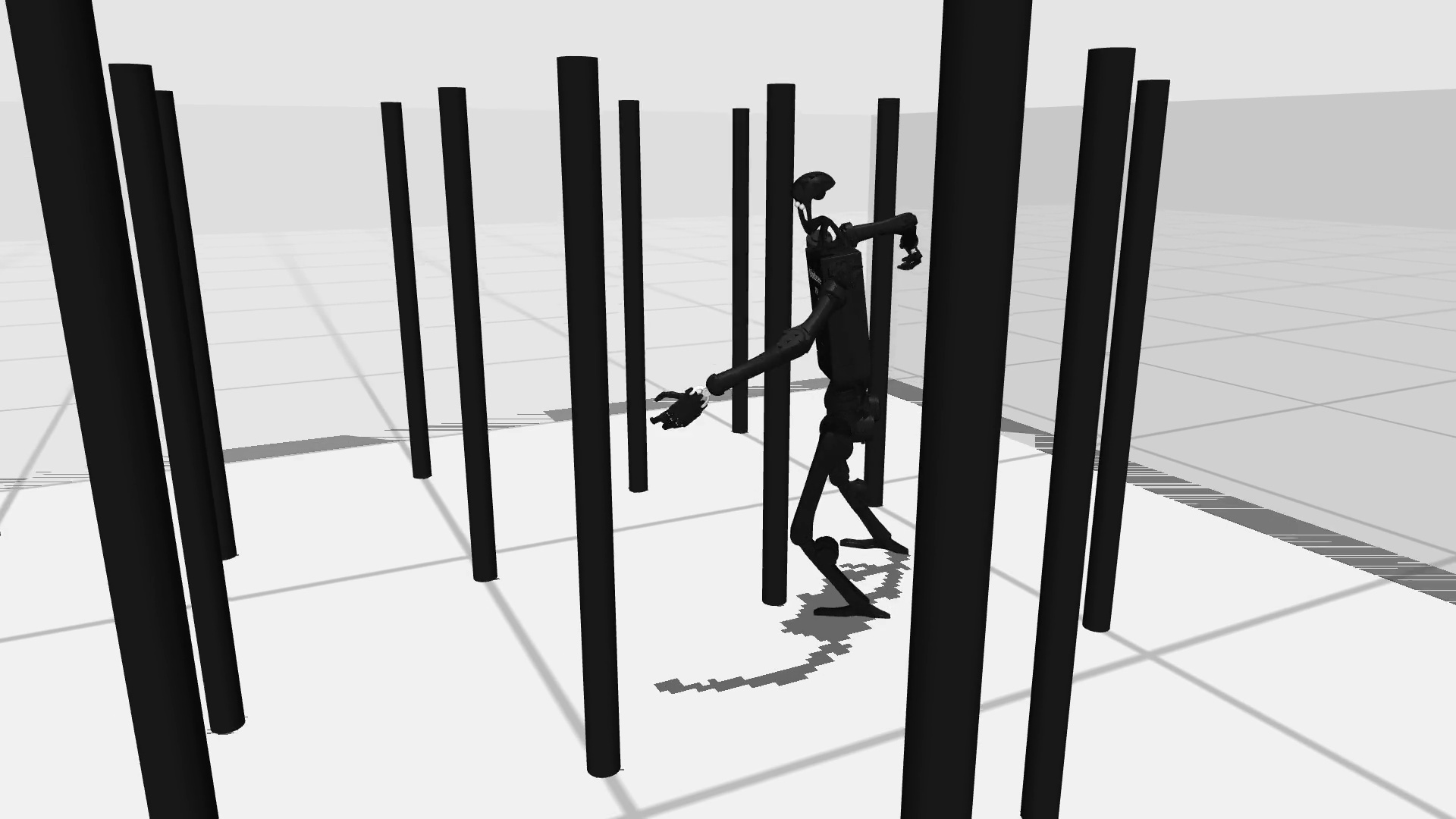}
    \caption{Frame 140}
  \end{subfigure}
  \begin{subfigure}[t]{0.242\textwidth}
    \centering
    \includegraphics[width=\linewidth]{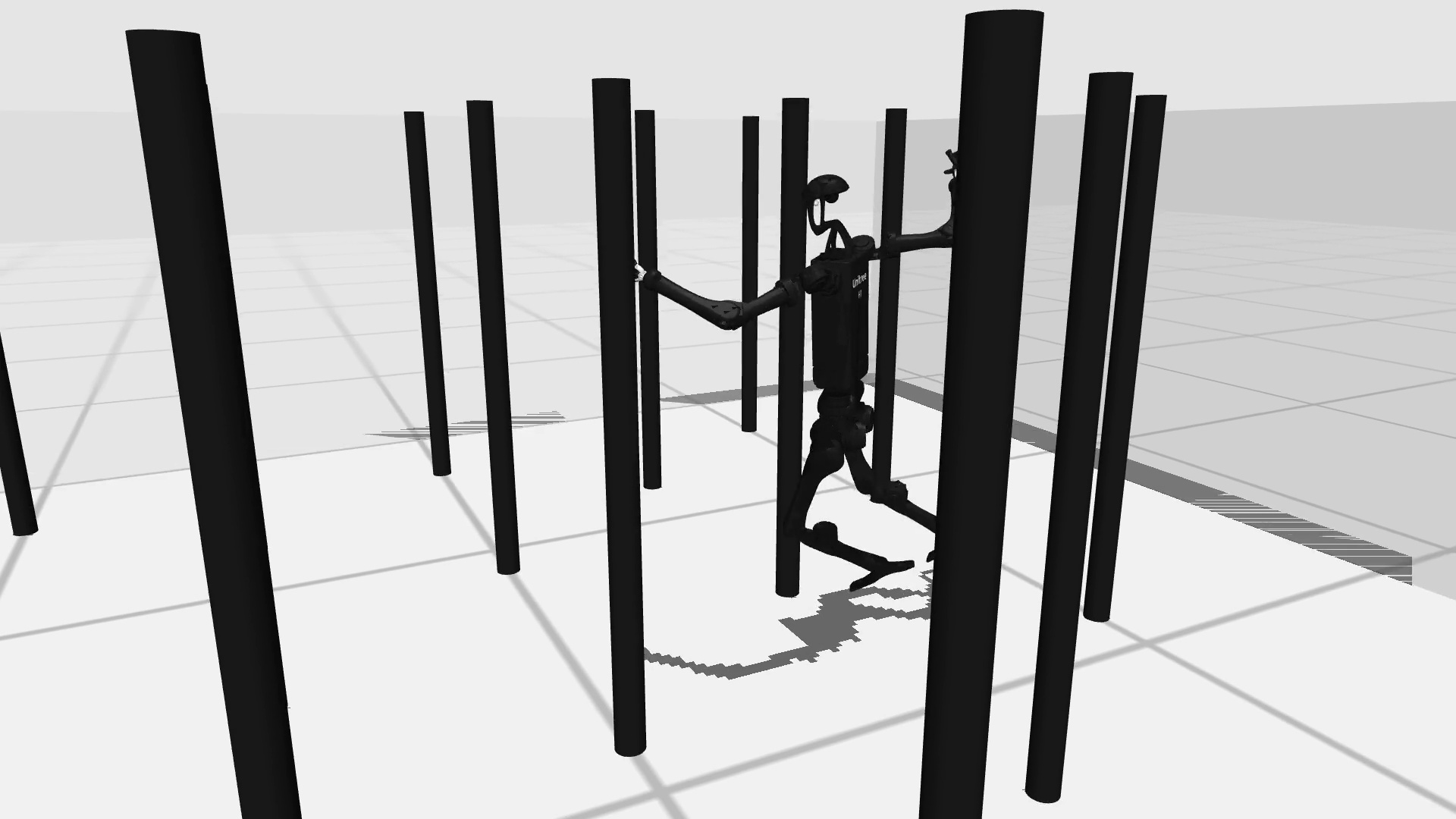}
    \caption{Frame 150}
  \end{subfigure}
  \begin{subfigure}[t]{0.242\textwidth}
    \centering
    \includegraphics[width=\linewidth]{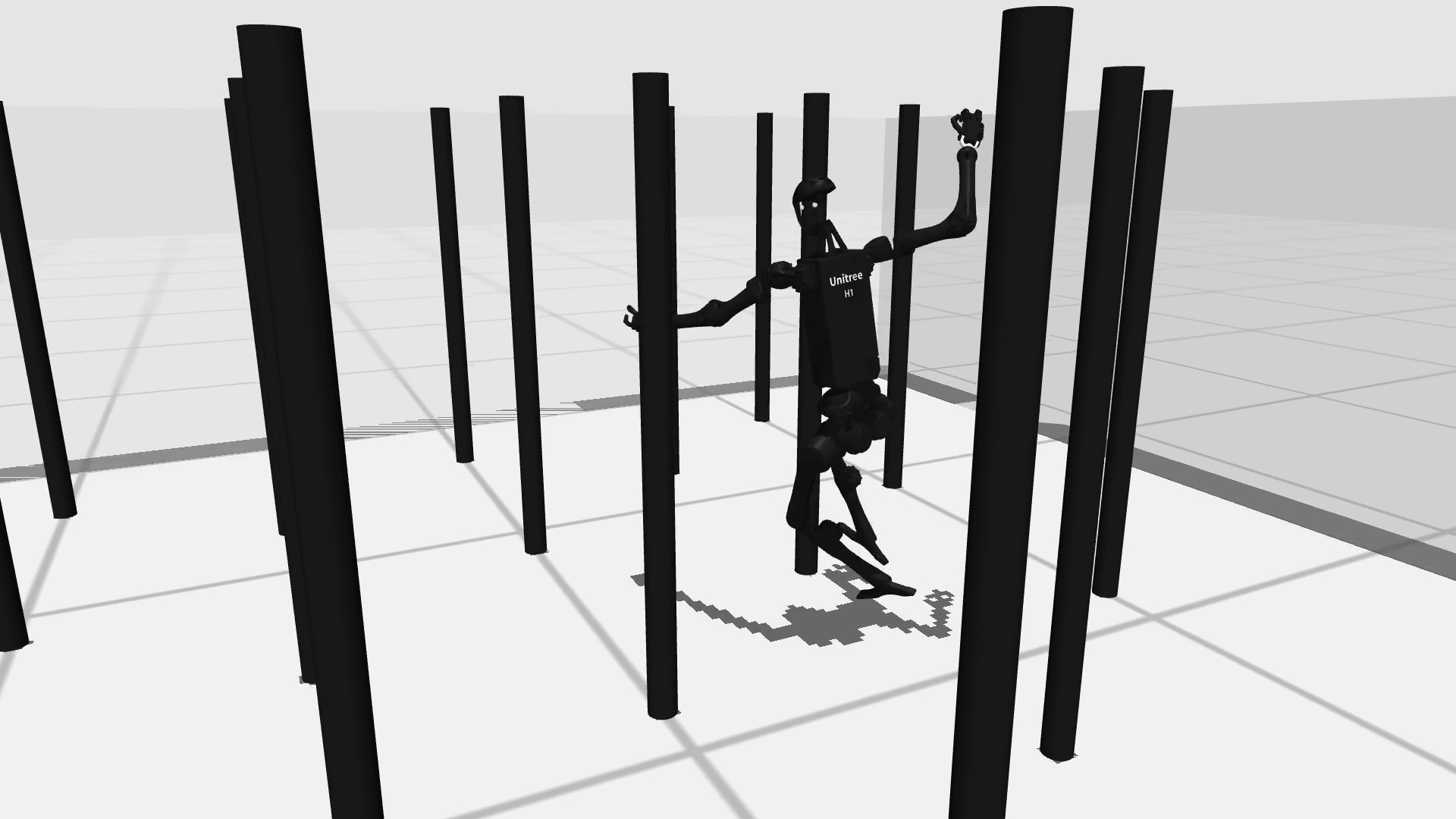}
    \caption{Frame 160}
  \end{subfigure}
  \begin{subfigure}[t]{0.242\textwidth}
    \centering
    \includegraphics[width=\linewidth]{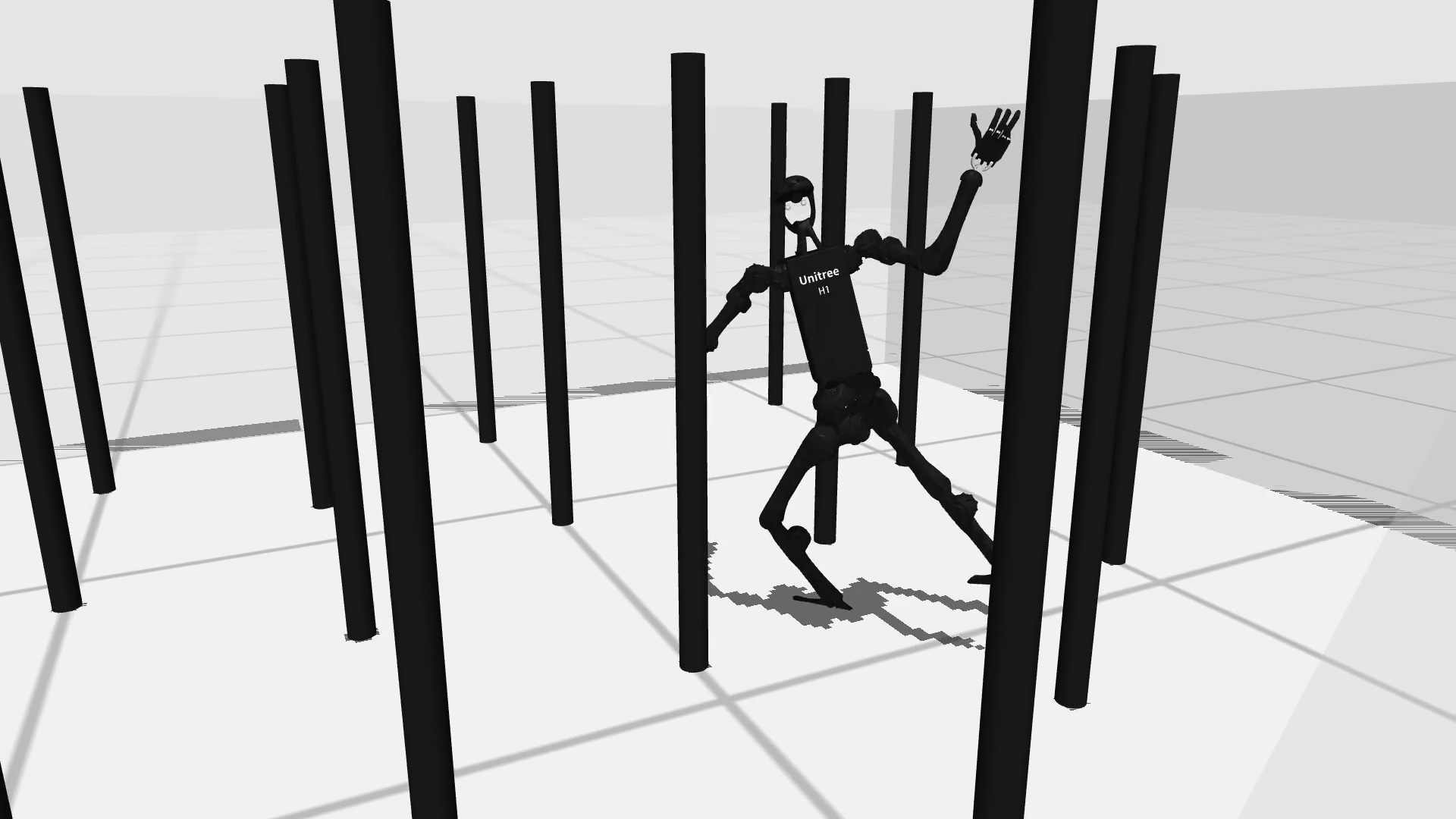}
    \caption{Frame 170}
  \end{subfigure}
    \begin{subfigure}[t]{0.242\textwidth}
    \centering
    \includegraphics[width=\linewidth]{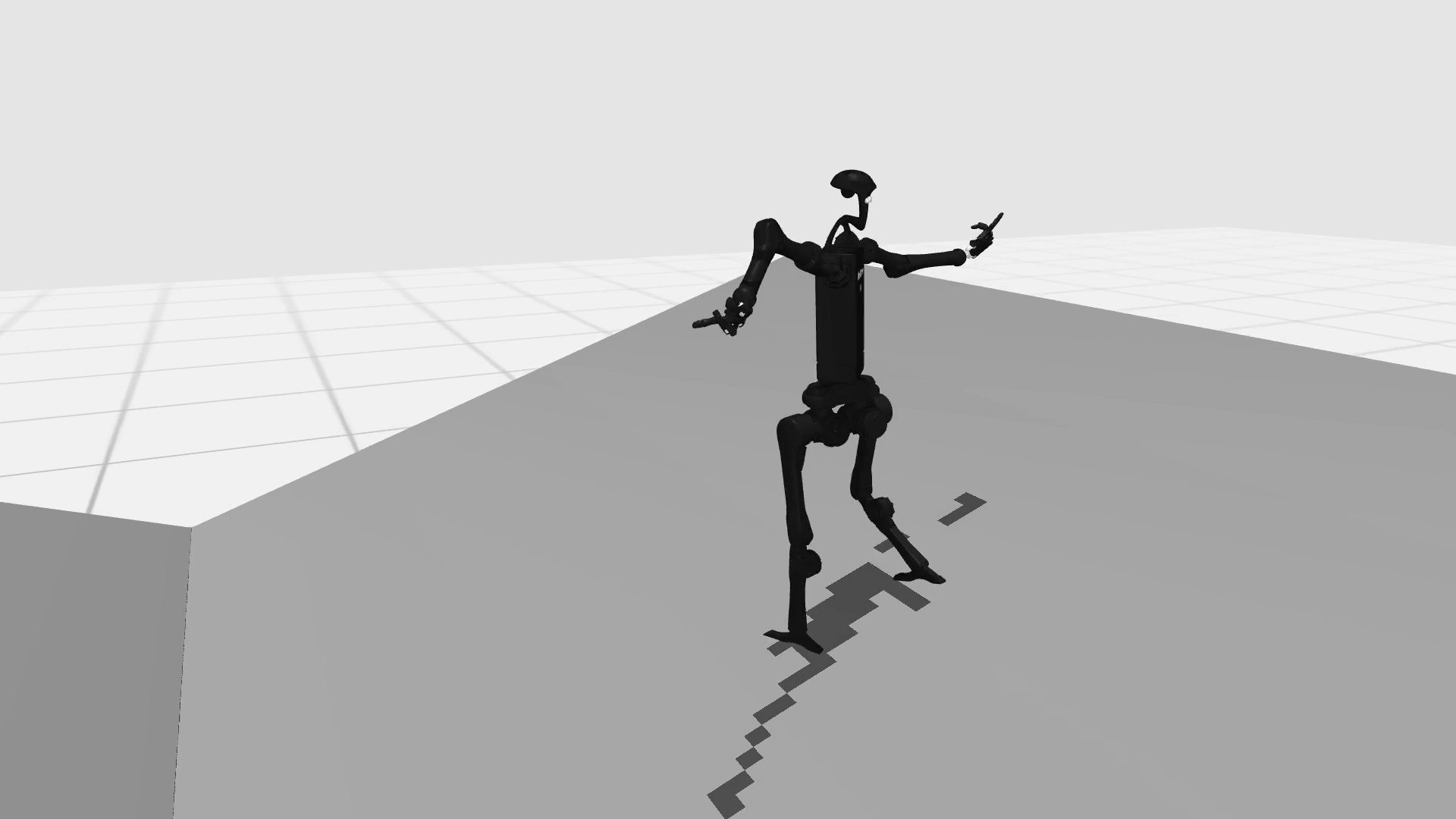}
    \caption{Frame 296}
  \end{subfigure}
  \begin{subfigure}[t]{0.242\textwidth}
    \centering
    \includegraphics[width=\linewidth]{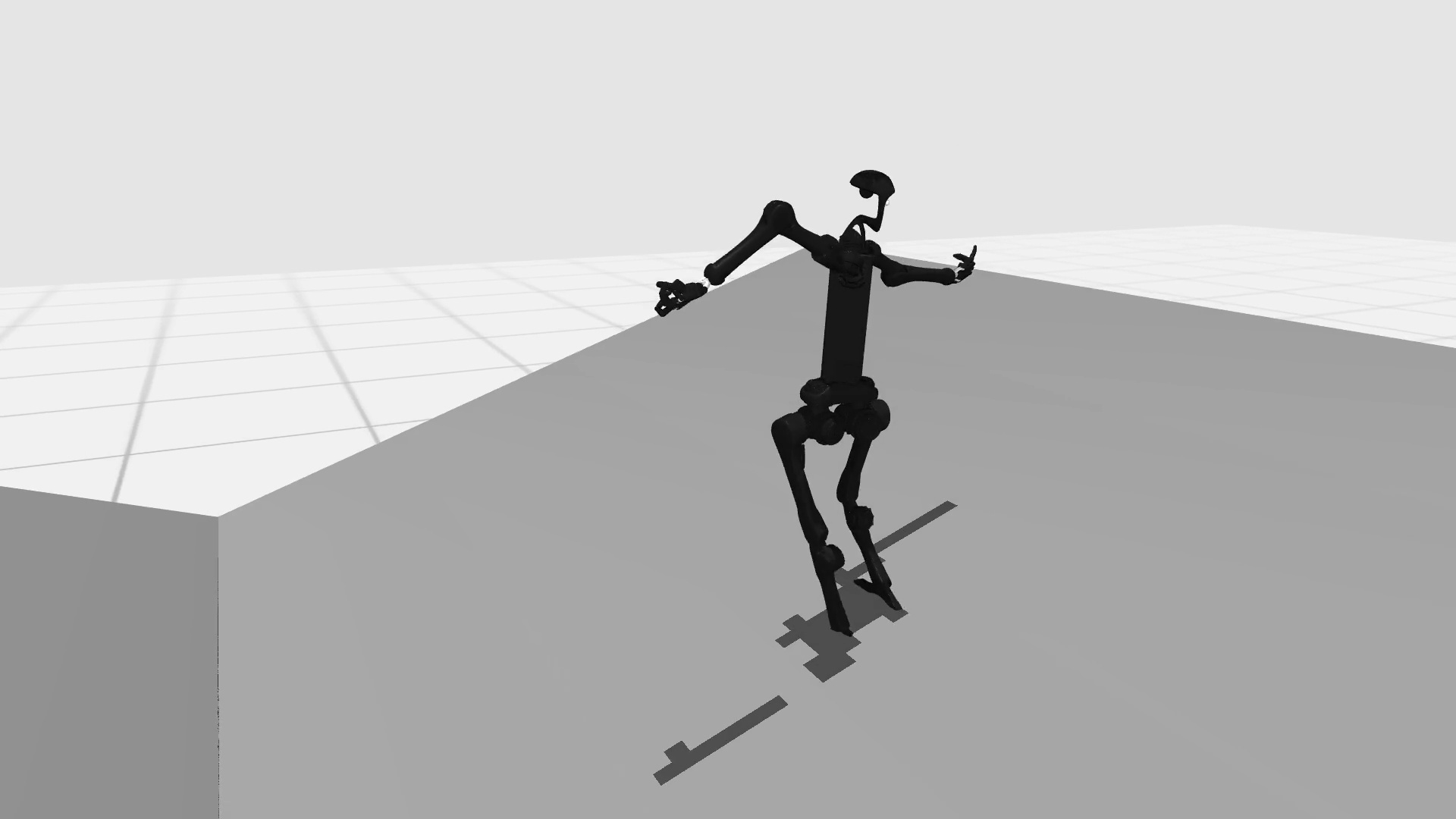}
    \caption{Frame 301}
  \end{subfigure}
  \begin{subfigure}[t]{0.242\textwidth}
    \centering
    \includegraphics[width=\linewidth]{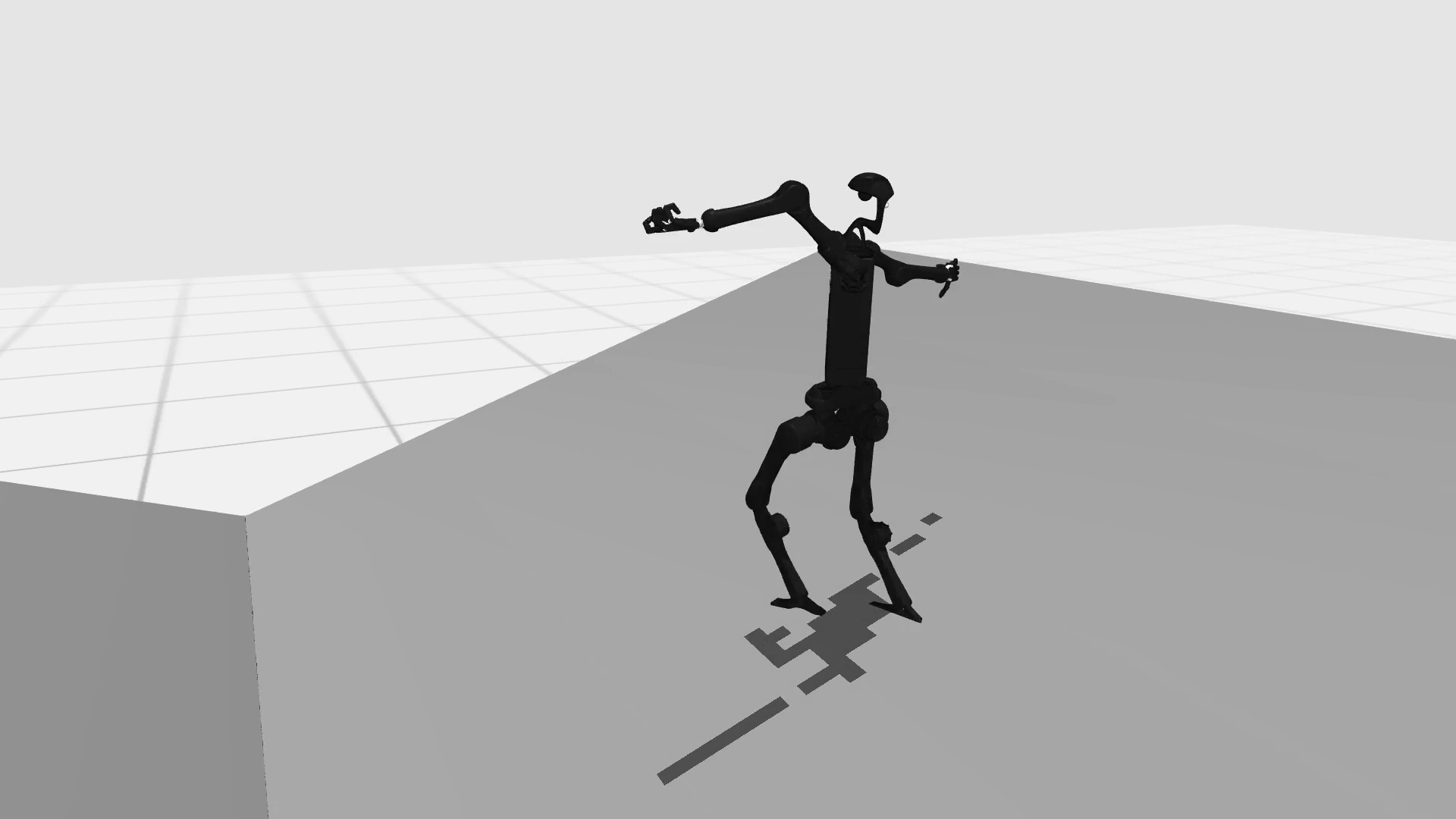}
    \caption{Frame 306}
  \end{subfigure}
  \begin{subfigure}[t]{0.242\textwidth}
    \centering
    \includegraphics[width=\linewidth]{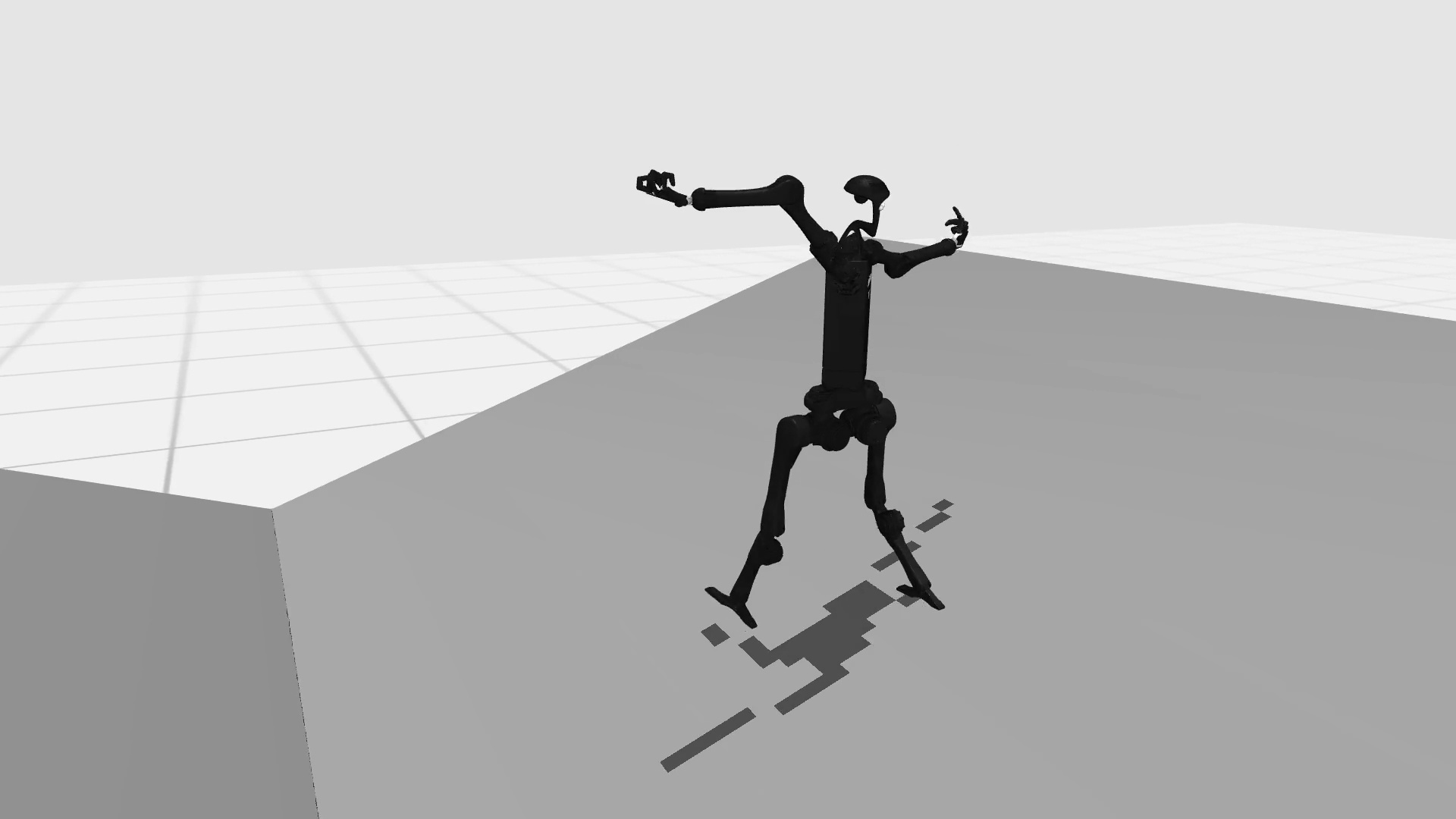}
    \caption{Frame 311}
  \end{subfigure}
  \caption{Visualizations of BOOM solving three of the most challenging tasks in the Humanoid Bench—\textbf{hurdle}, \textbf{pole}, and \textbf{slide}. }
  \label{fig:video_frames}
\end{figure}

\section{Limitation and Future Work}
\label{limit_future}
While BOOM demonstrates strong performance in high-dimensional continuous control, several limitations remain. First, the use of a planner during data collection introduces additional computational overhead, as generating and optimizing candidate actions through the world model is significantly slower than direct sampling in model-free methods. This can limit overall training throughput, particularly when environment interaction time is not the dominant bottleneck. Second, BOOM relies on a reasonably accurate dynamics model throughout training. When the model suffers from large prediction errors or distribution shift, especially over longer rollouts, both planning quality and value estimation may degrade, potentially harming overall performance.

Future work may explore accelerating the computation of online planning process and incorporating uncertainty-aware mechanisms to better handle multimodal or unreliable planner outputs. Additionally, extending BOOM to sparse-reward or real-world robotics settings with noisy observations also presents a promising and exciting direction.

\section{Positive and Negative Social Impact}
\label{impact}
Our method, BOOM, enhances the integration of online planning and off-policy learning in RL, leading to improved sample efficiency and final performance in high-dimensional control tasks. This has positive implications for real-world applications such as robotics and autonomous systems, where efficient and stable learning is crucial. By leveraging world models to reduce reliance on physical trials, our approach may also contribute to safer and more cost-effective training processes. However, like many RL technologies, BOOM could be misapplied in sensitive domains such as surveillance or autonomous weapons. Additionally, improved simulation-based efficiency might encourage premature deployment in safety-critical settings. We recommend cautious evaluation and responsible use to ensure the technology is applied ethically.


\newpage
\section*{NeurIPS Paper Checklist}

\begin{enumerate}

\item {\bf Claims}
    \item[] Question: Do the main claims made in the abstract and introduction accurately reflect the paper's contributions and scope?
    \item[] Answer: \answerYes{} 
    \item[] Justification: The abstract clearly outlines the key contributions and scope of the paper. It accurately presents (1) the motivation—divergence between planner and policy in planning-driven model-based RL, (2) the proposed method—BOOM, which tightly integrates planning and policy learning via a bootstrap loop, and (3) the novel techniques—likelihood-free divergence loss and soft value-weighted mechanism. It also specifies the evaluation domains (DeepMind Control Suite and Humanoid-Bench) and the claimed results (state-of-the-art performance in sample efficiency and final return). These elements are consistent with the core contributions described in the main paper.

\item {\bf Limitations}
    \item[] Question: Does the paper discuss the limitations of the work performed by the authors?
    \item[] Answer: \answerYes{} 
    \item[] Justification: The paper explicitly discusses several limitations of the proposed approach in Appendix \ref{limit_future}. It acknowledges that the use of a planner during data collection increases computational overhead compared to model-free methods, potentially limiting training throughput. Additionally, the method’s reliance on a reasonably accurate dynamics model is noted as a limitation—model errors or distribution shift can degrade both planning and value estimation, affecting performance. These points demonstrate a clear and honest discussion of the method’s constraints.

\item {\bf Theory assumptions and proofs}
    \item[] Question: For each theoretical result, does the paper provide the full set of assumptions and a complete (and correct) proof?
    \item[] Answer: \answerYes{} 
    \item[] Justification: The paper provides two formal theorems that analyze the theoretical guarantees of Bootstrap Alignment. For each result, we clearly state the assumptions (e.g., bounded reward, discount factor, forward KL divergence bound, Lipschitz continuity of the Q-function), and the statements are mathematically precise. While the main proofs are deferred to the appendix, the presentation in the main paper outlines the core logic and implications of the theorems.

    \item {\bf Experimental result reproducibility}
    \item[] Question: Does the paper fully disclose all the information needed to reproduce the main experimental results of the paper to the extent that it affects the main claims and/or conclusions of the paper (regardless of whether the code and data are provided or not)?
    \item[] Answer: \answerYes{} 
    \item[] Justification: The paper includes all necessary details to reproduce the main experimental results: it provides comprehensive descriptions of the benchmarks in Appendix \ref{benchmark}, and full hyperparameter settings in \ref{reproduce}. Our core algorithm file is accessible at \url{https://anonymous.4open.science/r/NeurIPS_BOOM-C587}. These elements together ensure that readers can independently verify and reproduce the main claims and conclusions, satisfying the reproducibility criterion.

\item {\bf Open access to data and code}
    \item[] Question: Does the paper provide open access to the data and code, with sufficient instructions to faithfully reproduce the main experimental results, as described in supplemental material?
    \item[] Answer: \answerYes{} 
    \item[] Justification:  This paper provides training curves, and numerical results (Section \ref{sec_exp}). Additionally, it offers a public repository \url{https://github.com/molumitu/BOOM_MBRL} containing the core implementation of the proposed algorithm. These elements together ensure that readers can independently verify and reproduce the main claims and conclusions, satisfying the reproducibility criterion.

\item {\bf Experimental setting/details}
    \item[] Question: Does the paper specify all the training and test details (e.g., data splits, hyperparameters, how they were chosen, type of optimizer, etc.) necessary to understand the results?
    \item[] Answer: \answerYes{} 
    \item[] Justification: The paper specifies the training and test setup in detail, including benchmark environments (Appendix \ref{benchmark}), full hyperparameter settings (Appendix \ref{reproduce}), and training curves (Figure \ref{fig_training_curves}). It clearly outlines choices such as the type of optimizer used, learning rates, horizon lengths for planning, and network structures. Hyperparameter values and their selection process are disclosed, allowing readers to understand and contextualize the reported performance.

\item {\bf Experiment statistical significance}
    \item[] Question: Does the paper report error bars suitably and correctly defined or other appropriate information about the statistical significance of the experiments?
    \item[] Answer: \answerYes{} 
    \item[] Justification: The paper includes training curves with error bars that reflect the variability across 3 random seeds in Figure \ref{fig_training_curves}, which is standard and appropriate for RL benchmarks. It also reports mean and standard deviation values for final performance metrics in Table \ref{tab_tar}, providing a clear sense of the statistical reliability of the results. These practices offer sufficient information about the significance and robustness of the experimental findings.

\item {\bf Experiments compute resources}
    \item[] Question: For each experiment, does the paper provide sufficient information on the computer resources (type of compute workers, memory, time of execution) needed to reproduce the experiments?
    \item[] Answer: \answerYes{} 
    \item[] Justification: The paper provides sufficient information on the computational setup used for the experiments in Section \ref{sec_exp}, including the type of compute workers (e.g., GPU/CPU), memory specifications, and total training time.
    
\item {\bf Code of ethics}
    \item[] Question: Does the research conducted in the paper conform, in every respect, with the NeurIPS Code of Ethics \url{https://neurips.cc/public/EthicsGuidelines}?
    \item[] Answer: \answerYes{} 
    \item[] Justification: We made sure the code was anonymous \url{https://anonymous.4open.science/r/NeurIPS_BOOM-C587}.

\item {\bf Broader impacts}
    \item[] Question: Does the paper discuss both potential positive societal impacts and negative societal impacts of the work performed?
    \item[] Answer: \answerYes{} 
    \item[] Justification:  In Appendix \ref{impact}, we discuss the potential positive and negative social impacts
of our work.
    
\item {\bf Safeguards}
    \item[] Question: Does the paper describe safeguards that have been put in place for responsible release of data or models that have a high risk for misuse (e.g., pretrained language models, image generators, or scraped datasets)?
    \item[] Answer: \answerNA{} 
    \item[] Justification:

\item {\bf Licenses for existing assets}
    \item[] Question: Are the creators or original owners of assets (e.g., code, data, models), used in the paper, properly credited and are the license and terms of use explicitly mentioned and properly respected?
    \item[] Answer: \answerYes{} 
    \item[] Justification: The paper properly credits all external assets used, such as open-source repositories including Pytorch-RL, DreamerV3 and TD-MPC2, by citing the original authors (Appendix \ref{reproduce}). These assets are used in accordance with their licenses and terms of use. Proper attribution ensures ethical reuse of resources and acknowledges the contributions of prior work, fulfilling this requirement.

\item {\bf New assets}
    \item[] Question: Are new assets introduced in the paper well documented and is the documentation provided alongside the assets?
    \item[] Answer: \answerYes{} 
    \item[] Justification: The paper introduces new assets, including a core implementation file and demonstration videos on three challenging tasks (\url{https://anonymous.4open.science/r/NeurIPS_BOOM-C587}). These assets effectively showcase the method's capabilities and support the main experimental claims. Upon acceptance, we will release the full codebase along with additional demos.

\item {\bf Crowdsourcing and research with human subjects}
    \item[] Question: For crowdsourcing experiments and research with human subjects, does the paper include the full text of instructions given to participants and screenshots, if applicable, as well as details about compensation (if any)? 
    \item[] Answer: \answerNA{} 
    \item[] Justification: 

\item {\bf Institutional review board (IRB) approvals or equivalent for research with human subjects}
    \item[] Question: Does the paper describe potential risks incurred by study participants, whether such risks were disclosed to the subjects, and whether Institutional Review Board (IRB) approvals (or an equivalent approval/review based on the requirements of your country or institution) were obtained?
    \item[] Answer: \answerNA{} 
    \item[] Justification:

\item {\bf Declaration of LLM usage}
    \item[] Question: Does the paper describe the usage of LLMs if it is an important, original, or non-standard component of the core methods in this research? Note that if the LLM is used only for writing, editing, or formatting purposes and does not impact the core methodology, scientific rigorousness, or originality of the research, declaration is not required.
    \item[] Answer: \answerNA{} 
    \item[] Justification:

\end{enumerate}

\end{document}